%% file: paper.tex
\documentclass[pdflatex,sn-mathphys]{sn-jnl}

\jyear{2023}%

\theoremstyle{thmstyleone}%
\usepackage{anyfontsize}

\theoremstyle{thmstyletwo}%
\usepackage{csquotes}
\usepackage[nameinlink]{cleveref}
\usepackage{capt-of}

\usepackage{subfigure}
\usepackage{verbatim}
\usepackage{amssymb}
\usepackage{comment}
\usepackage{acro}[=v2]
\usepackage[algo2e,ruled]{algorithm2e}
\theoremstyle{thmstylethree}%
\newtheorem{definition}{Definition}
\input{acronyms}
\begin{document}

\title[Interpretable predictions for patient pathways]{A machine learning framework for interpretable predictions in patient pathways: The case of predicting ICU admission for patients with symptoms of sepsis}


\author*[1]{\fnm{Sandra} \sur{Zilker}}\email{sandra.zilker@th-nuernberg.de}

\author[2]{\fnm{Sven} \sur{Weinzierl}}\email{sven.weinzierl@fau.de}

\author[3]{\fnm{Mathias} \sur{Kraus}}\email{mathias.kraus@informatik.uni-regensburg.de}

\author[4]{\fnm{Patrick} \sur{Zschech}}\email{patrick.zschech@uni-leipzig.de}

\author[2]{\fnm{Martin} \sur{Matzner}}\email{martin.matzner@fau.de}

\affil*[1]{\orgdiv{Technische Hochschule Nürnberg Georg Simon Ohm}, \orgname{Professorship for Business Analytics}, \orgaddress{\street{Hohfederstraße 40}, \city{90489 Nuremberg}, \country{Germany}}}

\affil[2]{\orgdiv{Friedrich-Alexander-Universität Erlangen-Nürnberg}, \orgname{Chair of Digital Industrial Service Systems}, \orgaddress{\street{Fürther Straße 248}, \city{90429 Nuremberg}, \country{Germany}}}

\affil[3]{\orgdiv{University of Regensburg}, \orgname{Chair for Explainable AI in Business Value Creation}, \orgaddress{\street{Bajuwarenstraße 4}, \city{93053 Regensburg}, \country{Germany}}}

\affil[4]{\orgdiv{Leipzig University}, \orgname{Professorship for Intelligent Information Systems and Processes}, \orgaddress{\street{Grimmaische Straße 12}, \city{04109 Leipzig}, \country{Germany}}}


\abstract{Proactive analysis of patient pathways helps healthcare providers anticipate treatment-related risks, identify outcomes, and allocate resources. Machine learning (ML) can leverage a patient's complete health history to make informed decisions about future events. However, previous work has mostly relied on so-called black-box models, which are unintelligible to humans, making it difficult for clinicians to apply such models. Our work introduces PatWay-Net, an ML framework designed for interpretable predictions of admission to the intensive care unit~(ICU) for patients with symptoms of sepsis. We propose a novel type of recurrent neural network and combine it with multi-layer perceptrons to process the patient pathways and produce predictive yet interpretable results. We demonstrate its utility through a comprehensive dashboard that visualizes patient health trajectories, predictive outcomes, and associated risks. Our evaluation includes both predictive performance -- where PatWay-Net outperforms standard models such as decision trees, random forests, and gradient-boosted decision trees -- and clinical utility, validated through structured interviews with clinicians. By providing improved predictive accuracy along with interpretable and actionable insights, PatWay-Net serves as a valuable tool for healthcare decision support in the critical case of patients with symptoms of sepsis.}



\maketitle

\newpage


\section*{Highlights}
\begin{itemize}
    \item This article proposes PatWay-Net, a novel machine learning framework for predicting critical pathways of patients with sepsis symptoms. Our framework retains patient pathway data in its natural form by combining non-linear multi-layer perceptrons (MLPs) for each static feature (i.e., static module) and an interpretable LSTM (iLSTM) cell for sequential features (i.e., sequential module). 
    \item Our results reveal that our approach outperforms commonly used interpretable machine learning models in our case, such as decision tree and logistic regression by 10.4\% and 7.3\% in terms of the area under receiver operating characteristic curve, respectively, and non-interpretable models, such as random forest and XGBoost by 4.4\% and 1.2\%, respectively. 
    \item PatWay-Net provides decision support to clinicians and hospital management in predicting the pathway of a patient accurately while remaining interpretable and can, therefore, help to improve hospital resource management.
    \item To enhance the model's interpretability and utility for clinical decision-makers, we have developed a comprehensive dashboard that visualizes patient health trajectories, predictive outcomes, and associated risks, facilitating informed clinical and resource allocation decisions.  
    \item The clinical utility of our framework is supported by structured interviews with independent clinicians, confirming its interpretability and actionable insights for healthcare decision support. 
\end{itemize}

\newpage

\section{Introduction}
\label{sec:intro}
As healthcare organizations face increasing demands and limited resources, the efficiency and compliance of healthcare processes are becoming increasingly important \citep{rojas2016}. The pandemic has served as a stress test for these processes, revealing several weaknesses, such as gaps in resource allocation, inefficiencies in patient triage, and limitations in data-driven decision-making \citep{morton2021introduction, bertsimas2021predictions}. As a remedy, advanced decision support systems based on modern \ac{ml} models can be employed to improve the performance of healthcare processes and provide proactive insights for clinical decision-makers~\citep{bertsimas2021predictions, janiesch_2021, kraus2020deep}. By using large amounts of data that are ubiquitously generated in today's healthcare information systems, such models can learn non-trivial patterns from historical patient trajectories.

A rich source of historical patient data is represented by so-called patient pathways, a timeline of each patient that describes the different departments, measurements, treatments, and transitions that a patient has gone through during a clinical stay \citep{barrera_ferro_improving_2020}. This information can be used to make accurate predictions about future health outcomes, informing the allocation of resources or the focus of medical professionals on specific patients \citep[e.g.,][]{barrera_ferro_improving_2020, yang_predicting_2010, elitzur2023machine, kramer2019classification}. In this way, healthcare institutions can derive recommendations for managing and controlling patient pathways early and identify risks and issues before they emerge.

Such recommendations are especially crucial in the context of sepsis symptoms, a complex and time-sensitive condition that demands rapid identification and intervention to improve patient outcomes \citep{komorowski2018artificial}. By leveraging patient pathway data, healthcare institutions can not only derive timely recommendations to manage and control disease progression but also identify risks and issues, such as early signs of sepsis, before they escalate \citep{mannhardt2017}. Consequently, early detection and treatment of sepsis, facilitated by the analysis of patient pathways, can significantly reduce a patient's deterioration.

\ac{ml} models represent a promising choice for predicting patient pathways as they can rapidly process large amounts of patient data and find latent patterns that help make informed decisions about patient outcomes. \ac{ml} models come in various forms and facets. For critical applications, clinical decision-makers typically favor interpretable\footnote{We make a strict distinction between the terms \textquote{interpretation} and \textquote{explanation}. Interpretation is derived from models designed to be intrinsically interpretable, whereas an explanation can be created by applying a post-hoc analytical explainable-artificial-intelligence approach to a black-box model (cf. \Cref*{sec:background}).} \ac{ml} models like decision trees, linear and logistic regression, and \acp{gam} \citep[e.g.,][]{lee_prediction_2020, Caruana.2015, Shipe2019, Bertoncelli.2020, magunia_machine_2021, bertsimas_optimal_2022}. They have the advantage of providing a clear understanding of how predictions are derived, which is crucial for making informed and accountable decisions. At the same time, however, such interpretable models have the limitation that they cannot handle sequential data structures in their natural form, limiting their prediction capabilities for time-varying patient data.

In contrast, there is an increasing interest in using more advanced and flexible models, such as bagged and boosted decision trees \citep[e.g.,][]{elitzur2023machine, barrera_ferro_improving_2020, kramer2019classification, liu_explainable_2023} or \acp{dnn} \citep[e.g.,][]{reddy_predicting_2018, ye_predicting_2020, zilker.2023}. \acp{dnn} are of particular interest for predicting patient pathways because of their ability to automatically discover and learn complex patterns in high-dimensional data \citep{lecun.2015}. This ability also allows them to capture hidden patterns in sequential data structures that are difficult to identify with traditional \ac{ml} models. However, \acp{dnn} generally have the limitation that they lack model interpretability because their internal decision logic is not directly comprehensible by humans \citep{janiesch_2021, BarredoArrieta2020}. This renders them black boxes for model developers and decision-makers, which is why they are unsuitable for critical healthcare applications.

To address the limitations of both research streams above, we propose PatWay-Net, an innovative \ac{ml} framework that is designed for both high predictive accuracy and intrinsic interpretability in modeling pathways from patients with symptoms of sepsis. With this framework, we leverage the principle of interpretable \ac{ml} models while harnessing the flexibility of a \ac{dnn} architecture. More specifically, our contributions are as follows:

\begin{itemize}
\item PatWay-Net is designed to constrain feature interactions, ensuring full model interpretability across the entire \ac{dnn} architecture.
\item The architecture blends non-linear \acp{mlp} for static features with an interpretable LSTM (iLSTM) cell for sequential features, preserving the natural data structure of patient pathways.
\item A comprehensive dashboard supports PatWay-Net's applicability by enabling clinical decision-makers to interpret PatWay-Net's predictive outcomes and associated risks easily.
\item Structured interviews with independent medical experts rigorously validate PatWay-Net's utility and interpretability, attesting to its real-world healthcare applicability.
\end{itemize}

We evaluate our proposed model using a real-life data set from an emergency department of a Dutch hospital, containing health records of patients with sepsis symptoms \citep{mannhardt2017}. During their stay, patients go through different activities (e.g., changing departments, receiving medications) and develop different trajectories of severity, resulting in individual patient pathways. The data set contains a rich set of static and sequential features, such as socio-demographic data, blood measurements, medical treatments, and diagnoses, which provide a valuable basis for predicting the future behavior of individual pathways. Specifically, we use this information to predict whether a patient will be admitted to the \ac{icu}, which constitutes a highly relevant prediction task for clinical professionals and administrative staff to support proactive resource allocation \citep{moerer2007german, lee_prediction_2020, kramer2019classification}. By comparing different types of \ac{ml} models, we show that PatWay-Net outperforms commonly used interpretable models, such as decision trees or logistic regression, and even non-interpretable models, such as random forest and XGBoost, in terms of area under the receiver operating characteristic curve ($AUC_{ROC}$) and F1-score. We then use PatWay-Net for interpreting both static and sequential features of the real-world setting to demonstrate its applicability for healthcare decision support.

Our paper is organized as follows: \Cref*{sec:clinical-relevance} motivates the task of predicting critical patient pathways from a clinical point of view. \Cref*{sec:background} presents relevant background and related work. \Cref*{sec:method} introduces our proposed \ac{ml} framework for interpretable patient pathway prediction, PatWay-Net. \Cref*{sec:eval} outlines the evaluation and application results based on the real-life use case for predicting \ac{icu} admission for patients with symptoms of sepsis. \Cref*{sec:discussion} summarizes our work by drawing implications for research and practice, reflecting on limitations, and providing an outlook for future work. 

\section{Clinical relevance}
\label{sec:clinical-relevance}

\subsection{Patient pathways and clinical decision support}
Healthcare processes are generally concerned with all activities related to diagnosing, treating, and preventing diseases to improve well-being~\citep{mans2015}. This includes patient-related activities organized in patient pathways and administrative activities that support clinical tasks~\citep{rebuge_2012}. Patient pathways are directly linked to a patient's diagnostic–therapeutic cycle and, therefore, do not constitute strictly standardized processes. However, accurate prediction of patient pathways is crucial for optimizing resource allocation, improving patient outcomes, and facilitating timely clinical interventions, thus making it an essential tool for enhancing healthcare efficiency and effectiveness.

\begin{figure}[ht]
\centering
	\includegraphics[width=0.9\textwidth]{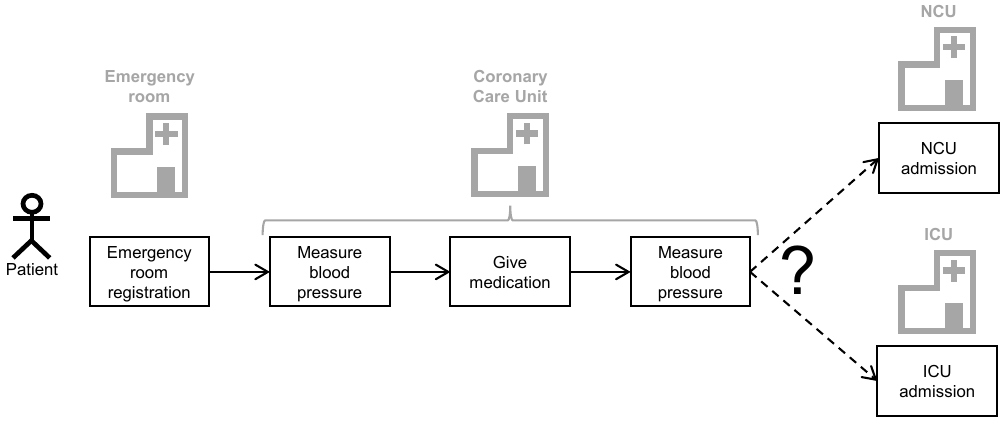} 
	\caption{Illustration of underlying setting. Multiple tasks must be performed when a patient is transferred to a new department or receives a new treatment. Thus, early prediction of the various steps a patient goes through during their hospital stay leads to more efficient operations.}
	\label{fig:setting}
\end{figure}

\Cref*{fig:setting} illustrates a patient's hospital stay at multiple departments. In each department, various tasks must be performed to ensure a safe and well-organized patient transition. In this example, the patient was transferred from the emergency room to the coronary care unit. Depending on the patient’s condition, the patient may be transferred to the \ac{icu} or the \ac{ncu}. Therefore, both departments must be prepared for patients. By using a decision support system that accurately predicts the next station, resources for one of the departments can be saved.

Technically, a patient visiting the hospital produces a patient pathway. A set of multiple patient pathways is then stored as an event log. \Cref*{tab:example} presents an example representation of an event log. Here, one visit of a patient is represented by a patient pathway (ID = 1). In the beginning, the patient registered at the emergency room at \emph{2024-02-20 12:11:01}. Also, the gender of the patient is registered as male (M). In the next patient activity, the blood pressure is measured at 180. Later, medication is administered before the blood pressure is measured for a second time at 195. The next activity then describes the patient being transferred to the \ac{icu}.

\begin{table}[ht]
\centering
\caption{Example event log with a single patient pathway following the scenario in \Cref*{fig:setting}.}
\label{tab:example}
\resizebox{\textwidth}{!}{
\begin{tabular}{@{}clccc@{}}
\toprule
\begin{tabular}[c]{@{}l@{}}Patient \\ pathway ID\end{tabular} & Patient activity & Timestamp & Blood pressure & Gender \\ \midrule
1 & Emergency room registration & 2024-02-20 12:11:01 & - & M \\
1 & Measure blood pressure & 2024-02-20 13:11:27 & 180 & M \\
1 & Give medication & 2024-02-20 14:30:27 & - & M \\
1 & Measure blood pressure & 2024-02-20 15:45:55 & 195 & M \\
1 & \ac{icu} Admission & 2024-02-20 16:12:02 & - & M \\
\bottomrule
\end{tabular}
}
\end{table}

As shown in \Cref*{tab:example}, the patient information in an event log is not structured to be easily processed by prediction models. Therefore, careful processing of static and sequential patient information is necessary to predict following patient activities accurately. In addition, experts generally have to ensure that the model learns meaningful patterns from the data, which constrains the model to be intrinsically interpretable. Both points are addressed in this work.

In the development of decision support for healthcare applications, the involvement of medical experts is inevitable \citep{wiens2019no}. Their insights can ensure that the proposed approach aligns with the complexities and problems of clinical practice. In this work, a comprehensive dashboard serves as a translational interface, bridging the gap between high-level computational outputs and real-world clinical decisions. It provides a demonstration of a model’s potential for real-world applicability, ensuring that its capabilities are both understandable and useful to healthcare practitioners.

\subsection{The case of sepsis}
\label{sec:clinical-relevance_sepsis}

Sepsis results from the body's overwhelming response to an infection and can be life-threatening \citep{rhodes2017surviving}. Therefore, sepsis is a time-sensitive issue that needs clinicians' attention as early as possible to enable the best possible outcome for each patient \citep{komorowski2018artificial}. Based on this, it is critical to predict this outcome during an ongoing patient pathway to provide timely recommendations for controlling the disease's progression \citep{mannhardt2017, rello2017sepsis}. However, the importance of sepsis lies not only in the urgency of its treatment but also in its complex and variable nature that can be detected in the resulting patient pathways \citep{schuurman2023embracing}. While its symptoms and, thus, underlying medical indicators, can progress or change rapidly, treatment needs to be adapted dynamically which influences the patient pathway \citep{rello2017sepsis}. Ultimately, in the context of developing interpretable \ac{ml} models for predicting patient pathways, the focus on patients with sepsis symptoms is crucial, given the imperative to enhance clinical decision-making, resource allocation, and ultimately, patient outcomes in this high-stakes domain.

\section{Methodological background and related work}
\label{sec:background}

\ac{ml} models are increasingly being integrated into clinical applications to assist healthcare professionals in diagnosing diseases, predicting patient outcomes, and making treatment decisions \citep{elitzur2023machine, kramer2019classification, yang_predicting_2010}. While the predictive power of these models is often decisive, it is also essential that they provide comprehensible outputs due to the critical nature of healthcare decisions. Comprehensible outputs promote transparency, reduce the risk of unintended biases, and ensure the reliability of the model results, ultimately contributing to safer and more effective patient care \citep{thorsen2020dynamic, loh_application_2022, rudin2019stop}. 

From a methodological point of view, there are generally two distinct streams of research dealing with comprehending \ac{ml} models. \Cref*{tab:context} provides an overview of both streams with exemplary approaches, which can be further classified according to the type of input features they support.

\begin{table}[ht]
\centering
\caption{Positioning of our work with respect to related fields from a methodological perspective.}
\footnotesize
\label{tab:context}
\makebox[\textwidth][c]{
\begin{tabular}{p{1.2cm}p{4.5cm}p{4.9cm}}
\toprule
& \textbf{Explainable machine learning} & \textbf{Interpretable machine learning} \\
\midrule
Definition & Refers to methods that aim to simplify (approximate) the decision logic of ML models that are not directly understandable to human users (known as black-box models).
& 
Refers to ML models that are designed to be inherently understandable to human users.

\\[1.5em]

Main focus & Encourages the use of flexible ML models with high predictive power that require post-hoc explanations to convert complex mathematical functions into a more understandable form for clinical model validation.
&
Encourages the use of ML models that ensure a complete understanding and validation of the decision logic for fully transparent clinical decision support without the need for additional explanation methods.
\\[1.5em]
Static features & 
Involves a scenario where a flexible black-box \ac{ml} model is provided with a fixed-length feature vector (e.g., age, weight, vital signs of a patient), and the model's response is analyzed after prediction using model-specific explanation methods such as layer-wise relevance propagation \cite[e.g.,][]{barrera_ferro_improving_2020} or model-agnostic explanation methods such as Shapley additive explanations \citep[e.g.,][]{liu_explainable_2023}.
& 
Interpretable \ac{ml} models limit interactions between features to reduce complexity, allowing for comprehensive validation of the model's performance. Typical interpretable \ac{ml} models are linear models \citep[e.g.,][]{lee_prediction_2020, Shipe2019, Bertoncelli.2020}, decision trees \citep[e.g.,][]{bertsimas_optimal_2022}, or generalized additive models \citep[e.g.,][]{Caruana.2015, magunia_machine_2021}, as well as typical risk charts, such as the well-known Simplified Acute Physiology Score (SAPS) at the intensive care unit~\citep{moreno_saps_2005}.\\[1.5em]

Sequential features & 
A black-box sequential \ac{ml} model is provided with temporal patient data (e.g., a trend of vital signs over a period), and the model's response is analyzed post-hoc \citep[e.g.,][]{thorsen2020dynamic,galanti2020explainable,Weinzierl2020XNAP}. Typical sequential \ac{ml} models with high predictive power are recurrent neural networks like long short-term memory (LSTM) networks \citep[e.g.,][]{ye_predicting_2020, reddy_predicting_2018}.
& 
An interpretable \ac{ml} model that processes temporal patient data with full transparency to medical professionals. Interpretable models allow for complete validation of model behavior. Examples include probabilistic finite automatons \citep[e.g.,][]{breuker.2016}, hidden Markov models \citep[e.g.,][]{lakshmanan_2015}, and certain advances in neural networks \citep[e.g.,][]{kaji2019attention, zhang2021interpretable}.
\\[1.5em]
Static + sequential features &
A black-box model consists of two parts: One that can process static features and one that can process sequential features \citep[e.g.,][]{zilker.2023, esteban2016predicting}. The information about the patient from the two sources is then combined to compute the model output. 
&
\textit{Our research:} A fully interpretable \ac{ml} model that can process both static and time-varying patient data.
\\
\bottomrule
\end{tabular}
}
\end{table}

\subsection{Explainable machine learning}
\label{sec:background_xml}

The first stream of research refers to the concept of explainable \ac{ml}. It promotes the use of flexible \ac{ml} models with high predictive power, which subsequently require \emph{post-hoc explanation methods} to convert their complex mathematical functions into easier-to-understand explanations \citep{BarredoArrieta2020, loh_application_2022}. Common representatives of flexible \ac{ml} models for static features are bagged and boosted decision trees such as random forest \citep{breiman2001random} and XGBoost \citep{chen2016xgboost}. Such models excel at handling static tabular data because they can capture complex interactions between features, allowing them to achieve high predictive performance \citep{barrera_ferro_improving_2020, elitzur2023machine, kramer2019classification, liu_explainable_2023}. In this work, we include both models as strong baseline approaches in our evaluation section. However, the construction of high-level interactions creates a lack of transparency because the individual feature effects are no longer understandable by humans and therefore require additional explanation methods.

For sequential features, the field has increasingly focused on \acp{dnn} in recent years \citep{lecun.2015}. Their multi-layered network architecture allows them to automatically discover and learn complex patterns in high-dimensional data structures that are relevant for the prediction task \citep{janiesch_2021, kraus2020deep}. Of particular interest are recurrent neural networks and \ac{lstm} networks because they can capture temporal patterns and therefore offer superior predictive performance compared to traditional approaches in dynamic and complex healthcare process environments~\citep[e.g.,][]{reddy_predicting_2018, ye_predicting_2020}. Furthermore, such network architectures have the advantage that they can be modified to capture static and sequential features simultaneously \citep[e.g.,][]{esteban2016predicting, zilker.2023}. Nevertheless, the nested, multi-layered structure of \acp{dnn} also creates a lack of transparency, because it is not directly observable what information in the input data drives the models to generate their prediction, rendering them black boxes for model users. In our work, we adopt the overall idea of an \ac{lstm} network \citep{hochreiter.1997} but propose a modification to ensure full model transparency.

To turn the internal decision logic of black-box models into comprehensible results, the field of explainable \ac{ml} has proposed a variety of post-hoc explanation methods \citep{barrera_ferro_improving_2020, rai_explainable_2020}. Some of these methods are model-specific. That is, they are designed for specific types of models and derive explanations by examining internal model structures and parameters (e.g., layer-wise relevance propagation for \acp{dnn} \citep{Weinzierl2020XNAP, barrera_ferro_improving_2020}). Other methods are model-agnostic and, therefore, broadly applicable to different \ac{ml} models. One of the most widely used model-agnostic methods is \ac{shap} \citep{lundberg.2017}. \ac{shap} uses a game-theoretic approach to explain the output of any \ac{ml} model. It has been applied, for example, to mortality prediction in \acp{icu} \citep{thorsen2020dynamic} and to process prediction models based on general event logs \citep{galanti2020explainable}. An overview of existing post-hoc explanation methods is given by Loh et al.~\citep{loh_application_2022}. Overall, post-hoc explanation methods have the advantage of providing a high degree of flexibility while encouraging the use of models with high predictive performance. Furthermore, they can lead to valuable insights, especially for exploratory analysis purposes \citep{senoner_using_2022}.

However, post-hoc explanation methods must also be viewed with caution. They generally attempt to reconstruct the cause of a generated prediction by approximation. As a result, they can never fully explain the entire black-box model without losing information, which may lead to unreliable results. Similarly, explanations are provided only \emph{after} a model's prediction, making it impossible to fully validate the functioning of the model for all inputs before model deployment. This issue becomes particularly critical when the distribution of input data changes over time, and the model may need to handle input feature ranges that were not encountered during its training phase. Overall, such deficiencies can lead to misleading conclusions and potentially harmful results~\citep{rudin2019stop, babic_beware_2021}. For this reason, we refrain from pursuing this general research stream in this paper.

\subsection{Interpretable machine learning}
\label{background_iml}

The second stream of research refers to the field of interpretable \ac{ml}, which promotes the development of \emph{intrinsically interpretable models} \citep{zschech_game_2022, BarredoArrieta2020}. In this research stream, the structure of an \ac{ml} model is constrained, such that the resulting model allows for a better understanding of how predictions are generated. Traditional representatives are linear models and decision trees, which are easy to comprehend and therefore often remain the preferred choice in critical healthcare applications \citep[e.g.,][]{lee_prediction_2020, Shipe2019, Bertoncelli.2020, bertsimas_optimal_2022}. At the same time, however, they are generally too restricted to capture more complex relationships.

A more advanced class of intrinsically interpretable \ac{ml} models are \acp{gam}~\citep{zschech_game_2022, BarredoArrieta2020}. In \acp{gam}, input features are modeled independently in a non-linear way to generate univariate shape functions that can capture arbitrary patterns but remain fully interpretable. The resulting shape functions for each feature are summed up afterward to produce the final model output. Thus, \acp{gam} include additive model constraints yet drop the linearity constraint of a simple logistic/ linear regression model. This structure is simply interpretable as it allows users to verify the importance of each feature. That is, the fitted shape functions directly reveal how each feature affects the predicted output without the need for additional explanation. 

In recent years, a wide variety of \ac{gam} variants have been proposed that can learn specific types of shape functions depending on the underlying learning procedure, for example, based on splines \citep{hastie_generalized_1986}, decision trees~\citep{lou_accurate_2013, lou_intelligible_2012}, or even neural networks \citep{yang2021gami, agarwal2021neural, kraus_interpretable_2023}. However, all of these approaches have in common that they primarily focus on processing static features and, therefore, cannot handle sequential data structures in their natural form \citep{zschech_game_2022}. As a consequence, their application in the healthcare domain is usually limited to preprocessed features in a static and aggregated form \citep[e.g.,][]{Caruana.2015, magunia_machine_2021}. In this work, we adopt the general idea of \acp{gam} to capture non-linear effects of individual features and propagate this idea not only to static features but also to sequential features to obtain a powerful yet fully interpretable model.

Apart from that, there are also interpretable \ac{ml} models that are specifically designed to capture sequential patterns. Traditional approaches include probabilistic finite automatons~\citep{breuker.2016} or hidden Markov models~\citep{lakshmanan_2015}. Such models have the drawback that they require explicit knowledge about the form of an underlying process model \citep{marquez-chamorro2018}, which is challenging to discover or reconstruct from complex event data in dynamic healthcare environments~\citep{mannhardt2017, HUANG2013111}. Therefore, recent approaches increasingly pursue the idea of constraining the structure of \ac{dnn} architectures to obtain models that can process sequential features in their natural form while remaining intrinsically interpretable. To date, however, little work exists in this area and current approaches often do not distinguish between sequential and static features \citep[e.g.,][]{kaji2019attention,zhang2021interpretable}.

In summary, only a limited amount of approaches deal with the development of intrinsically interpretable models for transparent patient pathway prediction. In particular, it lacks an innovative approach that can capture non-linear relationships in the form of flexible shape functions for static as well as sequential patient features while providing comprehensible model outputs that visualize the different feature effects for transparent decision support. Likewise, to the best of our knowledge, none of the existing approaches can automatically detect and integrate (sequential) feature interactions to control the model's flexibility for improved predictive performance. As a remedy, we propose PatWay-Net, a novel \ac{ml} framework that combines all these aspects within a single approach.

\section{PatWay-Net}
\label{sec:method}

This section describes PatWay-Net, an interpretable \ac{ml} framework building on a \ac{dnn} model with an architecture that transfers the ideas of \acp{gam} into a novel, intrinsically interpretable \ac{lstm} module for sequential features, and intrinsically interpretable \acp{mlp}, for static features.\footnote{For reproducibility, all developed and used material can be found here: \url{https://github.com/fau-is/patway-net}} We apply this proposed \ac{dnn} architecture of PatWay-Net to the problem of patient pathway prediction but want to emphasize that our proposed architecture is universal and can be applied to a variety of problem sets that combine sequential and static data~(see also Appendix~\ref*{app:furtherdata} for evaluations on other use cases).

In the following, we first describe the underlying problem of patient pathway prediction (\Cref*{sec:patwaynet_problem}), before mathematically describing the architecture (\Cref*{sec:patwaynet_architecture}) and the training process (\Cref*{sec:patwaynet_optimization}) of PatWay-Net's \ac{dnn} model. Subsequently, we describe the different interpretation plots that can be derived from the intrinsically interpretable architectural design of PatWay-Net's \ac{dnn} model (\Cref*{sec:patwaynet_interpretation}). 

\subsection{Problem statement}
\label{sec:patwaynet_problem}

An \ac{ml} model $f \in \mathcal{F}$ should map patient pathways to a target of interest, with~$\mathcal{F}$ denoting the so-called hypothesis space. Patient pathways comprise two sets of information, one set describes static information about the patient, and one set describes dynamic or sequential information about the patient.

\begin{definition}[Patient Pathways]
Mathematically, the information that describes a set of patients can be expressed as a tuple 
\begin{equation}
\left(\mathbf{X}_{static}, \, \mathsf{X}_{seq}\right),
\end{equation}
where $\mathbf{X}_{static} \in \mathbb{R}^{s\times q}$ is the static patient data, and $\mathsf{X}_{seq} \in \mathbb{R}^{s\times T \times p}$ is the sequential patient data. The \mbox{dimension $s$} denotes the number of patient pathways, $q$ indicates the number of static variables that describe a patient (e.g., one-time diagnoses or gender), and $T$ and $p$ describe the number of time steps that we recorded for the sequential information and the number of features tracked in each time step, respectively. A single patient's patient pathway $i$ is denoted by the static information $\mathbf{X}_{static}^{(i)}$ and the sequential data $\mathsf{X}_{seq}^{(i)}$. 

\end{definition}

The objective of this work is to find a prediction model $f \in \mathcal{F}$ that maps the static information $\mathbf{X}_{static}$ and sequential information $\mathsf{X}_{seq}$ about patients to target outcomes $\mathbf{y} = (y_{1}, \dots, y_{s})$, that is
\begin{equation}
    f: \left(\mathbf{X}_{static}, \, \mathsf{X}_{seq}\right) \rightarrow \mathbf{y}.
\end{equation}
The target outcomes $\mathbf{y}$ can thereby represent various patient activities in the future, such as \ac{icu} admission. 

A timely prediction of the future occurrence of an activity is crucial, as it can prevent the worsening of the patient’s condition and initiate successful treatment by medical experts. Therefore, a prediction model should not only make predictions once the full patient pathway is present but should make predictions already at earlier stages, that is, with less information included in the patient pathways. Thus, we define the patient pathway prefix in the following. 

\begin{definition}[Patient Pathway Prefix]
Given patient pathway $i$ with static information $\mathbf{X}_{static}^{(i)} \in \mathbb{R}^{q}$ and sequential information $\mathsf{X}_{seq}^{(i)} \in \mathbb{R}^{T \times p}$, the patient pathway prefix of length $t^*$ is defined as a tuple
\begin{equation}
 \left(\mathbf{X}_{static}^{(i)}, \, \mathsf{X}_{seq}^{(i)}[:t^*]\right),
\end{equation}
where $\mathsf{X}_{seq}^{(i)}[:t^*] \in \mathbb{R}^{t^* \times p}$ denotes the first $t^*$ time steps of the patient's sequential information. 
\end{definition}

\subsection{Architecture of the DNN model}
\label{sec:patwaynet_architecture}

\begin{figure}[ht]
\centering
	\includegraphics[width=\textwidth]{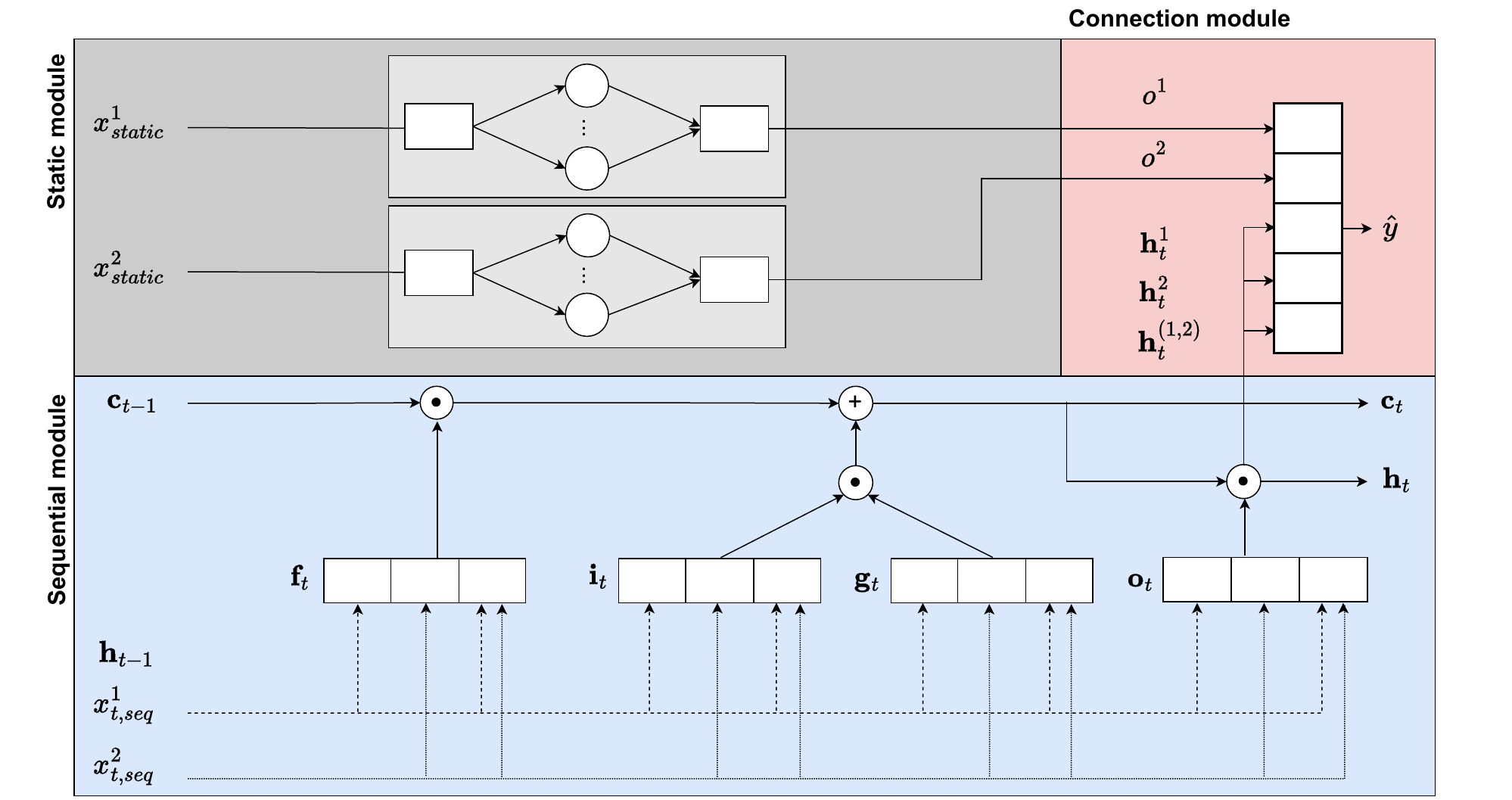} 
	\caption{Illustration of the architecture of PatWay-Net's \ac{dnn} model consisting of a sequential, a static, and a connection module. Here, two static and two sequential features are shown, which run through their modules and are then connected.}
	\label{fig:architecture} 
\end{figure}

The proposed interpretable architecture of PatWay-Net is shown in \Cref*{fig:architecture}. It contains a static, a sequential, and a connection module. While the first two modules naturally model the event log data, the connection module maps the outputs of these modules onto predictions of patient activities (in our case \ac{icu} admission).

\subsubsection{Static module}

The static module resembles a \ac{gam} \citep{hastie_generalized_1986}, yet combines the underlying idea with the power of \acp{dnn} \citep{agarwal2021neural}. By making this architectural choice, we allow our proposed model to remain fully transparent. That is, the effect of each input feature on the model output can be fully assessed after training the model. This is achieved by mapping the input features separately to output values (i.e., there are no interactions between input features). This separation naturally constrains this \ac{dnn} but, on the other hand, allows the visual inspection of the effect each static feature has on the network's output. Consequently, although our proposed model is derived from the field of \acp{dnn}, we make careful choices about our architecture to allow for a fully transparent white-box model (in contrast to the black-box behavior of general \acp{dnn}).

Mathematically, for $q$ static input features $\mathbf{X}_{static}[1], \ldots, \mathbf{X}_{static}[q]$, the static module maps the input features to outputs $o^{1}, \ldots, o^{q}$ through 
\begin{equation}
    o^{l} = f_{MLP}^l(\mathbf{X}_{static}[l]), \qquad \text{with } l \in \{1,\ldots,q\},
\end{equation}
where each $f_{MLP}^l$ denotes a neural network and $o^{l}\in \mathbb{R}$ indicates a single scalar. The neural networks of the architecture are trained in individual sub-modules so that the weights of the different neural networks are trained independently from each other (cf. the boxes around the neural networks of the static module in \Cref*{fig:architecture}). With this architecture, we can later compute the outputs $o^l$ for various input values for each neural network $f_{MLP}^l$ and, thereby, visually inspect the effect that the input has on the output. 

\subsubsection{Sequential module}
The sequential module extends the previous idea of our static module to a sequential setting. For this, we propose a novel \ac{ilstm} layer to encode the values of each sequential feature $\mathsf{X}_{seq}[j]$ into a vector $\mathbf{h}_{t}^{j} \in \mathbb{R}^{m}$, with $j \in \{1,2, \dots, p\}$, where $m$ denotes the hidden size for a single sequential feature in the \ac{ilstm} cell. 
To ensure intrinsic interpretability of the \ac{ilstm}, each sequential feature has its corridor throughout the gates and state vectors of the original \ac{lstm}~\citep{hochreiter.1997}, without the possibility to interact with any other feature (similar to the previous static module, in which each static feature went through a separate neural network).
Such feature corridors in the \ac{ilstm} layer have a specific size, defined by the internal element size of the corresponding sequential feature $m$, defining how much vector space is reserved for the sequential feature value computation, from the gates to the hidden state. 

Similar to a vanilla \ac{lstm}~\citep{hochreiter.1997}, the \ac{ilstm} uses a forget gate, an input gate, and an output gate, as well as a candidate state, resulting in the vectors $\mathbf{f}_t$, $\mathbf{i}_t$, $\mathbf{o}_t$, and $\mathbf{\Tilde{c}}_t$, respectively. The information of the sequence is then stored in a cell state $\mathbf{c}_{t}$, and a hidden state $\mathbf{h}_{t}$. Technically, this restriction, to not allow uncontrolled interactions, is realized by multiplying weight matrices with masking matrices. A masking matrix includes only 0 or 1 values. If an element of a weight matrix should be considered, the corresponding element in the masking matrix is set to 1, else it has the value 0. Mathematically, the \ac{ilstm} can be formalized~as 
\begin{align}
\mathbf{f}_t &= \sigma\left(\mathbf{x}_{t} \otimes (\mathbf{U}_f * \mathbf{U}_m) + \mathbf{h}_t \otimes (\mathbf{V}_f * \mathbf{V}_m) + \mathbf{b}_f\right),\\
\mathbf{i}_t &= \sigma\left(\mathbf{x}_{t} \otimes (\mathbf{U}_i * \mathbf{U}_m) + \mathbf{h}_t \otimes (\mathbf{V}_i * \mathbf{V}_m) + \mathbf{b}_i\right),\\
\mathbf{o}_t &= \sigma\left(\mathbf{x}_{t} \otimes (\mathbf{U}_o * \mathbf{U}_m) + \mathbf{h}_t \otimes (\mathbf{V}_o * \mathbf{V}_m) + \mathbf{b}_o\right),\\
\mathbf{\Tilde{c}}_t &= tanh\left(\mathbf{x}_{t} \otimes (\mathbf{U}_{\Tilde{c}} * \mathbf{U}_m) + \mathbf{h}_t \otimes (\mathbf{V}_{\Tilde{c}} * \mathbf{V}_m) + \mathbf{b}_{\Tilde{c}}\right),\\
\mathbf{c}_{t+1} &= \mathbf{f}_{t} * \mathbf{c}_{t} + \mathbf{i}_{t} * \mathbf{\Tilde{c}}_{t}, \\
\mathbf{h}_{t+1} &= \mathbf{o}_{t} * tanh\left(\mathbf{c}_{t+1}\right).
\end{align}
Here, $\sigma$ denotes the sigmoid activation, $\otimes$ is the matrix multiplication, and~$*$ denotes the element-wise multiplication. $\mathbf{U}_m$ and $\mathbf{V}_m$ are masking matrices that ensure that the individual features are computed independently using values from their corridor and, therefore, omitting interactions between sequential features. 
By contrast, a traditional, non-interpretable \ac{lstm}~\citep{hochreiter.1997} does not use such masking matrices and, therefore, allows any interaction between features for which values are to be computed. 
As output, the \ac{ilstm} layer returns for each sequential feature $\mathsf{X}_{seq}[j]$ the vector $\textbf{h}^{j} \in \mathbb{R}^{m}$, that is, the last hidden state of the \ac{ilstm} for the sequential \mbox{feature $j$}. Let $f_{iLSTM}^j \in \mathbb{R} \rightarrow \mathbb{R}^m$ denote this function, which maps the $j$-th sequential feature onto the corresponding hidden state, and let $f_{iLSTM} \in \mathbb{R}^p \rightarrow \mathbb{R}^{p*m}$ denote the function that maps all sequential features to the complete hidden state vector.

Beyond single sequential features, the \ac{ilstm} layer can encode values of a pairwise sequential feature interaction $(j,k)$ in $(1,\ldots,p) \times (1,\ldots,p)$ into a vector $\mathbf{h}^{j,k} \in \mathbb{R}^{m}$. Mathematically, the \ac{ilstm} computes such interactions as an additional sequential feature that does not interact with other features. The interactions to be used in PatWay-Net can be chosen manually or can be detected automatically using heuristics. We describe such a heuristic in Appendix~\ref*{app:use_case}. 

\subsubsection{Connection module}

In the connection module, the information from the static module and the sequential module are then combined to compute the estimations $\mathbf{\hat{y}}$ for the target outcomes $\mathbf{y}$. Mathematically, we use the hidden outcome values $o^1,\ldots,o^q$ for static features, the hidden state values $\mathbf{h}^{1},\ldots,\mathbf{h}^{p}$ for sequential features, and potentially $\mathbf{h}^{j,k}$ for interacting sequential features $j,k\in(1,\ldots,p) \times (1,\ldots,p)$. These values are then concatenated and mapped onto the output neuron to provide the estimations $\mathbf{\hat{y}}$. The mapping is performed using a single feed-forward layer with sigmoid activation, as the prediction of patient activities (in our case \ac{icu} admission) is defined as a binary classification task. 

\subsection{Parameter optimization of the DNN model}
\label{sec:patwaynet_optimization}

All parameters from the three modules are combined into one \ac{dnn} model in which these are optimized simultaneously. Let $f_{\text{PatWay-Net}}$ denote this \ac{dnn} model with parameters $\beta$. Depending on the task, the fit of $f_{\text{PatWay-Net}}$ to the target outcomes $\mathbf{y}$ is then measured by a loss function $\mathcal{L}$. In our real-world data application, we use binary cross-entropy, as \ac{icu} admission represents a binary decision. Overall, we minimize the empirical risk, that is
\begin{align}
 \beta^* &=\underset{\beta}{\arg \min}\sum^{s}_{i=1}\sum^T_{t=1}\mathcal{L}\Big(f_{{\text{PatWay-Net}}}\Big(\mathbf{X}^{(i)}_{static},\mathsf{X}^{(i)}_{seq}[:\,t]; \beta\Big), y_i\Big),
\end{align}
where we iterate over the patient pathways $s$ and over the prefixes for each patient pathway~$T$.

We address this optimization problem using an adaptive moment estimation (Adam) optimizer \citep{KingBa15} with default hyperparameters.
For every epoch, we perform a mini-batch gradient descent to optimize the internal parameters batch-wise efficiently.

\subsection{Interpretations of the DNN model}
\label{sec:patwaynet_interpretation}

Based on the architectural design of PatWay-Net's \ac{dnn} model, different interpretation plots can be created, allowing an interpretation of how the model input affects the model output. The interpretation plots are part of a comprehensive dashboard, that serves as a decision support tool for clinical decision-makers (cf. \Cref*{sec:eval_interpretation}). \Cref*{tab:overview_plots} provides an overview of the four interpretation plots that we propose in this paper, including plot names, the underlying equations, and short descriptions of the plots' purposes.

In the real-life data application that follows, we prefer the designation \emph{(medical) indicator} over \emph{feature} because it is more comprehensible for decision-makers in the medical domain. Accordingly, we name our four interpretation plots \emph{medical indicator importance}, \emph{medical indicator shape}, \emph{medical indicator transition}, and \emph{medical indicator development} (see \Cref*{tab:overview_plots}). The importance, shape, and transition plots are based on so-called shape functions \citep{lou_intelligible_2012}. Traditionally, shape functions are only computed for static features \citep[e.g.,][]{Caruana.2015, magunia_machine_2021, yang2021gami, lou_intelligible_2012, lou_accurate_2013, agarwal2021neural, kraus_interpretable_2023}. However, one of our paper's contributions is that we also extend their idea to sequential features to obtain interpretable model results for sequential features.

\begin{table}[ht]
\centering
\caption{Overview of PatWay-Net's interpretation plots.}
\label{tab:overview_plots}
\resizebox{\textwidth}{!}{
\begin{tabular}{p{0.2\textwidth}p{0.15\textwidth}p{0.65\textwidth}}
\toprule
Plot name &Underlying equation& Description  \\ \midrule
Medical indicator importance & (\ref*{equ:global_static_feature_shape}) and (\ref*{equ:global_sequential_feature_shape})& This plot shows the medical indicator importance on the $x$-axis and the medical indicator name on the $y$-axis for all static and sequential data. It provides a quick overview of which medical indicators are most relevant for the model.  \\[0.5em]
Medical indicator shape & (\ref*{equ:global_static_feature_shape}) and (\ref*{equ:global_sequential_feature_shape}) & This plot shows the medical indicator value on the $x$-axis and the effect on model output on the $y$-axis for static (\ref*{equ:global_static_feature_shape}) or sequential (\ref*{equ:global_sequential_feature_shape}) data. It allows medical experts to get a detailed look into the model behavior for single points in time. \\[0.5em]
Medical indicator transition & (\ref*{equ:global_sequential_feature_transition}) & This plot shows the effect that a transition of a sequential medical indicator has from a value at time step $t-1$ to another value at time step $t$. It depicts the change of effect on the $z$-axis. Thereby, medical experts can observe how changes in indicator values over time (e.g., vital signs) affect the model. \\[0.5em]
Medical indicator development & (\ref*{equ:local_sequential_feature_effect}) & This plot shows time steps on the $x$-axis and respective effect values on the $y$-axis for sequential data. It provides the trajectory as well as the effect that each value had on the model.
\\
\bottomrule
\end{tabular}
}
\end{table}

In general, shape functions describe the effect on the model output for various values of a single indicator. Thus, these plots answer the question, \emph{\textquote{How does the model output change for various values of a medical indicator?}}. For a static indicator $l$, the shape function represents the function described by $f_{MLP}^l$ and the corresponding parameters in the connection module, that is, it shows values $\mathbf{X}_{static}[j]$ for the $l$-th static indicator on the x-axis and %
\begin{align}
\label{equ:global_static_feature_shape}
    f_{MLP}^l(\mathbf{X}_{static}[l]),
\end{align}
weighted by the parameters in the connection module, on the y-axis. For sequential data $j$, the \ac{ilstm} layer can be illustrated similarly, with the $x$-axis showing the values $\mathsf{X}^{(i)}_{seq}[j, t]$ of a sequential indicator $j$ for an individual pathway and a single time step $t$, and the $y$-axis, showing the corresponding effects on the model output via
\begin{align}
    \label{equ:global_sequential_feature_shape}
    f_{iLSTM}^j(\mathsf{X}^{(i)}_{seq}[j, :t]),
\end{align} 
weighted by the parameters in the connection module. This plot can also be extended to interactions of two sequential indicators with a three-dimensional plot, in which the color denotes the interaction effect on the model output, as exemplarily shown in Appendix~\ref*{app:use_case}. We call this plot the sequential medical indicator interaction plot. Note that the $f_{iLSTM}$ layer preserves the history up to time step $t$. In our case, we are showing the effect of sequential data, depending on the history of a patient's pathway.

An indicator importance can be derived by computing the area under the shape functions, that is, under the plots that are described in \Cref*{equ:global_static_feature_shape} and \Cref*{equ:global_sequential_feature_shape}. Thereby, PatWay-Net allows computing the overall importance of static and sequential indicators, answering the question, \emph{\textquote{Which medical indicators are the most important?}}. 

A medical indicator transition illustrates the change of effect on the model output for sequential data from time step $t-1$ to $t$. This answers the question, \emph{\textquote{How does the model output change from the previous to the current time step?}}. 
To answer this question, for a given sequential indicator $j$ with the sequence $\mathsf{X}^{(i)}_{seq}[j]$, we calculate the difference in the effect in the sequential module between the time steps $t$ and $t-1$, that is, we calculate
\begin{align}
    \label{equ:global_sequential_feature_transition}
    f_{iLSTM}^j(\mathsf{X}^{(i)}_{seq}[j, t]) - f_{iLSTM}^j(\mathsf{X}^{(i)}_{seq}[j, t-1]),
\end{align}
weighted by the parameters in the connection module. To illustrate all combinations of changes in the value from the last to the current time step, along with the change of effect on the model output, we use a three-dimensional plot, in which the $z$-axis (the color) describes an increase or decrease in the change of the effect.
 
Lastly, a medical indicator development describes the trajectory of an indicator over time. Due to the design of PatWay-Net's \ac{dnn} model, we can illustrate the sequential effect over time. That is, the model output can be tracked for each time step of the patient pathway's sequential information and plotted afterward. 
This plot is specifically useful to answer the question, \emph{\textquote{What effect did a sequential medical indicator of a given patient pathway have on the model output over time?}}. 
As such, we adopt the general idea of Weinzierl et al.~\cite{Weinzierl2020XNAP} to provide transparency at the local instance level. However, instead of using a post-hoc explanation method, we can directly plot the interpretable results from our sequential and connection module. Mathematically, this can be derived for a sequence $\mathsf{X}^{(i)}_{seq}[:\,t], \text{ } t=1,\ldots,T$ of indicator $j$ through a plot, showing the time steps $1,\ldots,T$ on the x-axis, and, on the y-axis,
\begin{align}
    \label{equ:local_sequential_feature_effect}
    f_{iLSTM}(\mathsf{X}^{(i)}_{seq}[j,:\,t]), \qquad \text{with } t = 1,\ldots,T,
\end{align}
weighted by a scalar value from the connection module. 

\section{Evaluation and application of PatWay-Net}
\label{sec:eval}

We evaluate PatWay-Net and demonstrate its applicability using a real-world use case from a Dutch hospital. After introducing the use case (\Cref*{sec:eval_case}) and describing the baseline models (\Cref*{sec:eval_baselines}), we perform a three-step evaluation procedure. First, we evaluate the predictive performance of PatWay-Net's intrinsically interpretable \ac{dnn} model through a benchmark study (\Cref*{sec:eval_pred_perf}). Second, we evaluate the meaningfulness of PatWay-Net's interpretation plots through a demonstration as part of a comprehensive dashboard for clinical decision-makers and a discussion including clinical evidence of the visualized interpretation aspects~(\Cref*{sec:eval_interpretation}). Finally, we validate the utility of PatWay-Net for decision-makers through structured interviews with clinicians from different hospitals and different domains. The results of those interviews are also presented in \Cref*{sec:eval_interpretation}, while additional information can be found in Appendix~\ref*{sec:eval_interviews}.\footnote{We further verify and demonstrate the validity of PatWay-Net's created interpretation plots by comparing these with post-hoc-generated explanations for black-box models, and verify the end-to-end training capability of PatWay-Net's \ac{dnn} model (Appendix~\ref*{app:use_case}). 
We also perform a simulation study (Appendix~\ref*{app:simulationstudy}) and verify the prediction capability of PatWay-Net's \ac{dnn} model with two additional data sets from other domains (Appendix~\ref*{app:furtherdata}).} 

\subsection{Use case description}
\label{sec:eval_case}
Our real-life publicly-available data set comprises pathways of patients with sepsis symptoms from a Dutch hospital with approximately 50,000 patients per year~\citep{mannhardt2017}.\footnote{\url{https://data.4tu.nl/articles/dataset/Sepsis_Cases_-_Event_Log/12707639}} The hospital uses an \ac{erp} system to track all performed patient events. The process consists of logistical activities, including the patient's stations through the hospital, and medical activities, such as blood value measurements and medical treatments. 
Although the aforementioned process can be described in a fairly structured manner based on the information provided by the use case provider~\citep{mannhardt2017}, this structure is only reflected to a limited extent in the underlying event log. In addition, patients can run through different activities in highly individual pathways, making it difficult to detect patterns to estimate an individual pathway’s outcome manually.

Based on these patient pathways, we predict whether a patient will be admitted to the \ac{icu}. This prediction is highly relevant for both healthcare providers and insurance companies. First, capacity and staff planning in \acp{icu} are crucial and influence the patient's probability of recovery~\citep{moreno1998nursing}. Second, admissions to the \ac{icu} for septic patients are among the highest costs compared to other diseases \citep{moerer2007german}.

The patient events of the event log can be differentiated into 16 activities with different purposes, for example, release type, type of measurement, or stating whether the patient was admitted to normal care. They all represent sequential medical indicators. In addition to the control-flow information, the event log contains another 27 indicators. Three are sequential and numerical and represent the measured values of \emph{\ac{crp}}, \emph{Leukocytes}, and \emph{LacticAcid}. Furthermore, patient \emph{Age} is a numerical and static indicator. The summary statistics of the numerical indicators are presented in \Cref*{tab:descripitveStatofNumVar}.

\begin{table}[ht]
\centering
\caption{Summary statistics of the numerical medical indicators in our real-life data set.}
\label{tab:descripitveStatofNumVar}
\begin{tabular}{lrrrrrrrr}
\toprule
\multirow{2}{*}{Medical indicator}    & \multirow{2}{*}{Obs.} &    \multirow{2}{*}{Mean}     &  \multirow{2}{*}{SD}         &  \multicolumn{5}{c}{Percentile} \\ \cmidrule{5-9}

   &       &        &         & 5\% & 25\% & 50\% & 75\% & 95\% \\ \midrule
Age        & 724   & 72.12  & 15.48   & 40.0  & 65.0   & 75.0   & 85.0   & 90.0   \\
\ac{crp}        & 2,388 & 111.66 & 83.53   & 12.0  & 44.0   & 94.0   & 156.0  & 276.0  \\
LacticAcid & 992   &  1.98 &1.49    & 0.7 & 1.1  & 1.6  & 2.3  & 4.7    \\
Leukocytes & 2,525 & 13.24  & 16.87  & 2.8 & 7.6  & 11.0 & 15.1 & 24.9 \\ \bottomrule
\multicolumn{9}{l}{\footnotesize{\emph{Note:} Obs. = Number of observations, SD = Standard deviation.}}
\end{tabular}

\end{table}

Besides \emph{Age}, there are 22 categorical static indicators (e.g., type of medical staff executing the activity), or binary values (e.g., stating whether or not the patient received an infusion)~\citep{mannhardt2017}. To avoid data leakage, we remove the medical indicator \emph{diagnosis} as the large majority of the patient pathways with certain diagnoses describe patients who are later admitted to the \ac{icu}. That is, it can be assumed that the hospital guidelines require all patients with a certain diagnosis to be admitted to the \ac{icu}. A detailed description of the further data preprocessing steps, as well as evidence that the size of the used event log is appropriate for PatWay-Net's \ac{dnn} model to achieve accurate and timely predictions, can be found in Appendix~\ref*{app:use_case}.

\subsection{Baseline models}
\label{sec:eval_baselines}

We benchmark PatWay-Net\footnote{For simplicity, we will sometimes refer to PatWay-Net's \ac{dnn} model as PatWay-Net in the following.} against three groups of \ac{ml} approaches. The first group includes decision tree, $K$-nearest neighbor, na\"ive Bayes, and logistic regression. This group allows us to assess how well PatWay-Net performs compared to traditional shallow \ac{ml} models that are intrinsically interpretable. These models are limited to processing static patient information and cannot handle sequential data. 
The second group includes random forest and XGBoost. This group allows us to assess how well PatWay-Net performs compared to commonly used black-box \ac{ml} approaches. Like the first group, these models are limited to processing static patient information and cannot handle sequential data.
Third, we include a state-of-the-art \ac{lstm} model that uses the static module of PatWay-Net and combines it with an unrestricted \ac{lstm} cell \citep{hochreiter.1997} to process the sequential patient information. This model represents the model with the highest flexibility and modeling capacity. Yet, it does not allow for a transparent interpretation of how the predictions are derived. Thus, its applicability in high-stakes decisions is generally limited.

These baseline models are then compared to our proposed PatWay-Net. Here, we compare two versions. First, PatWay-Net without any interactions between the sequential medical indicators. Second, PatWay-Net with pairwise interaction between a set of sequential medical indicators.\footnote{Information on further experiments with multiple pairwise interactions can be found in Appendix~\ref*{app:use_case}.} 
In doing so, we tune models by applying a grid search, evaluate models by performing a five-fold stratified cross-validation, and measure the predictive performance of models by calculating $AUC_{ROC}$ and F1-score. 
More details about our model tuning, model evaluation, and model selection can be found in Appendix~\ref*{app:use_case}. 

\subsection{Results on predictive performance}
\label{sec:eval_pred_perf}

The predictive performance of PatWay-Net and the baselines for the use case are summarized in \Cref*{tab:results}. Among the interpretable shallow \ac{ml} models, logistic regression and na\"ive Bayes models outperform the decision tree and $K$-nearest neighbor models with an improvement of 1.9 to 8.9 percentage points in terms of $AUC_{ROC}$ performance on the test sets.

PatWay-Net outperforms all shallow \ac{ml} baseline models across all metrics. We observe that PatWay-Net, without any interactions, pushes the predictive performance by 5.1\% in comparison to shallow \ac{ml} models. By incorporating an interaction term in our sequential module, we achieve an $AUC_{ROC}$ performance on the test sets of 0.734, which is an improvement of 7.3\% compared to the logistic regression, and an improvement of 10.4\% compared to the decision tree. PatWay-Net with the interaction in the sequential module even outperforms the non-interpretable models XGBoost and random forest by 4.4\% and 1.2\%, respectively.  
We conduct a Friedman test and a Wilcoxon signed-rank test with Holm p-value adjustment \citep{demsar2006}, which shows that the difference is statistically significant with $\alpha = 1\%$ for the decision tree, logistic regression, and $K$-nearest neighbor models, and with $\alpha = 10\%$ for the na\"ive Bayes model. Further information on the statistical tests can be found in Appendix~\ref*{app:use_case}.

As an upper bound, the state-of-the-art \ac{lstm} network leads to an $AUC_{ROC}$ performance on the test sets of 0.757, which is only slightly higher than our proposed PatWay-Net. However, it does not allow any intrinsic model interpretation.

\begin{table}[ht]
\centering
\caption{Comparison between baseline models and PatWay-Net. Highlighted are the best performances among interpretable models.}\label{tab:results}
\resizebox{\textwidth}{!}{
\begin{tabular}{@{}lrrlrr@{}}
\toprule
\begin{tabular}{l}\multirow{2}{*}{\ac{ml} approach}\end{tabular} & \multicolumn{2}{c}{F1-score (weighted)}                   &  & \multicolumn{2}{c}{$AUC_{ROC}$}                               \\ \cmidrule(lr){2-3} \cmidrule(l){5-6} 
                          & \multicolumn{1}{c}{Validation} & \multicolumn{1}{c}{Test} &  & \multicolumn{1}{c}{Validation} & \multicolumn{1}{c}{Test} \\ \midrule
\multicolumn{6}{l}{\footnotesize{\textsc{Our Approach}}}                                                                 \\
\begin{tabular}{l}\textbf{PatWay-Net}\\\textbf{(with interaction)}  \end{tabular}               & 0.886 \footnotesize{($\pm$.016)}       & \textbf{0.896} \footnotesize{($\pm$.016)} &  & 0.820 \footnotesize{($\pm$.028)}       & \textbf{0.734} \footnotesize{($\pm$.058)} \\
\begin{tabular}{l}\textbf{PatWay-Net}\\\textbf{(without interaction)}  \end{tabular}                & 0.883 \footnotesize{($\pm$.015)}       & 0.893 \footnotesize{($\pm$.016)} &  & \textbf{0.821} \footnotesize{($\pm$.027)}       & 0.724 \footnotesize{($\pm$.049)} \\
\multicolumn{6}{l}{\footnotesize{\textsc{Interpretable Shallow Machine Learning}}}                                                                                           \\
\begin{tabular}{l}Decision tree     \end{tabular}                   &     0.879 \footnotesize{($\pm$.019)}                             &     0.890 \footnotesize{($\pm$.016)}                       &  &            0.753 \footnotesize{($\pm$.060)}                      &    0.665 \footnotesize{($\pm$.069)}                        \\
\begin{tabular}{l}$K$-nearest neighbor  \end{tabular}                     &    \textbf{0.892} \footnotesize{($\pm$.019)}                             &      0.859 \footnotesize{($\pm$.025)}                     &  & 0.673 \footnotesize{($\pm$.047)}                                &        0.600 \footnotesize{($\pm$.049)}                   \\
\begin{tabular}{l}Na\"ive Bayes    \end{tabular}                    &  0.363 \footnotesize{($\pm$.242)}                               &       0.416 \footnotesize{($\pm$.228)}                    &  &      0.723 \footnotesize{($\pm$.043)}                           &          0.689 \footnotesize{($\pm$.056)}                 \\
\begin{tabular}{l}Logistic regression \end{tabular}                       &   0.881 \footnotesize{($\pm$.016)}                              &    0.890 \footnotesize{($\pm$.015)}                       &  &    0.769 \footnotesize{($\pm$.044)}                             &      0.684 \footnotesize{($\pm$.063)}                     \\
\multicolumn{6}{l}{\footnotesize{\textsc{Non-Interpretable Machine Learning}}}\\
\begin{tabular}{l}\ac{lstm} network\\(with static module) \end{tabular}                     &    0.890 \footnotesize{($\pm$.018)}                            &   0.898 \footnotesize{($\pm$.014)}                       &  &        0.840 \footnotesize{($\pm$.028)}                        &   0.757 \footnotesize{($\pm$.049)}                       \\
\begin{tabular}{l}XGBoost \end{tabular}                       &   0.883 \footnotesize{($\pm$.018)}                              &    0.896 \footnotesize{($\pm$.016)}   &  &    0.817 \footnotesize{($\pm$.014)}                             &      0.703 \footnotesize{($\pm$.018)}\\
\begin{tabular}{l}Random forest \end{tabular}                       &   0.881 \footnotesize{($\pm$.017)}                              &    0.885 \footnotesize{($\pm$.011)}   &  &    0.804 \footnotesize{($\pm$.013)}                             &      0.725 \footnotesize{($\pm$.016)}                     \\
\bottomrule
\end{tabular}
}
\end{table}

\subsection{Results on interpretation}
\label{sec:eval_interpretation}

\begin{figure}[ht]
    \centering
    \includegraphics[width=\columnwidth]{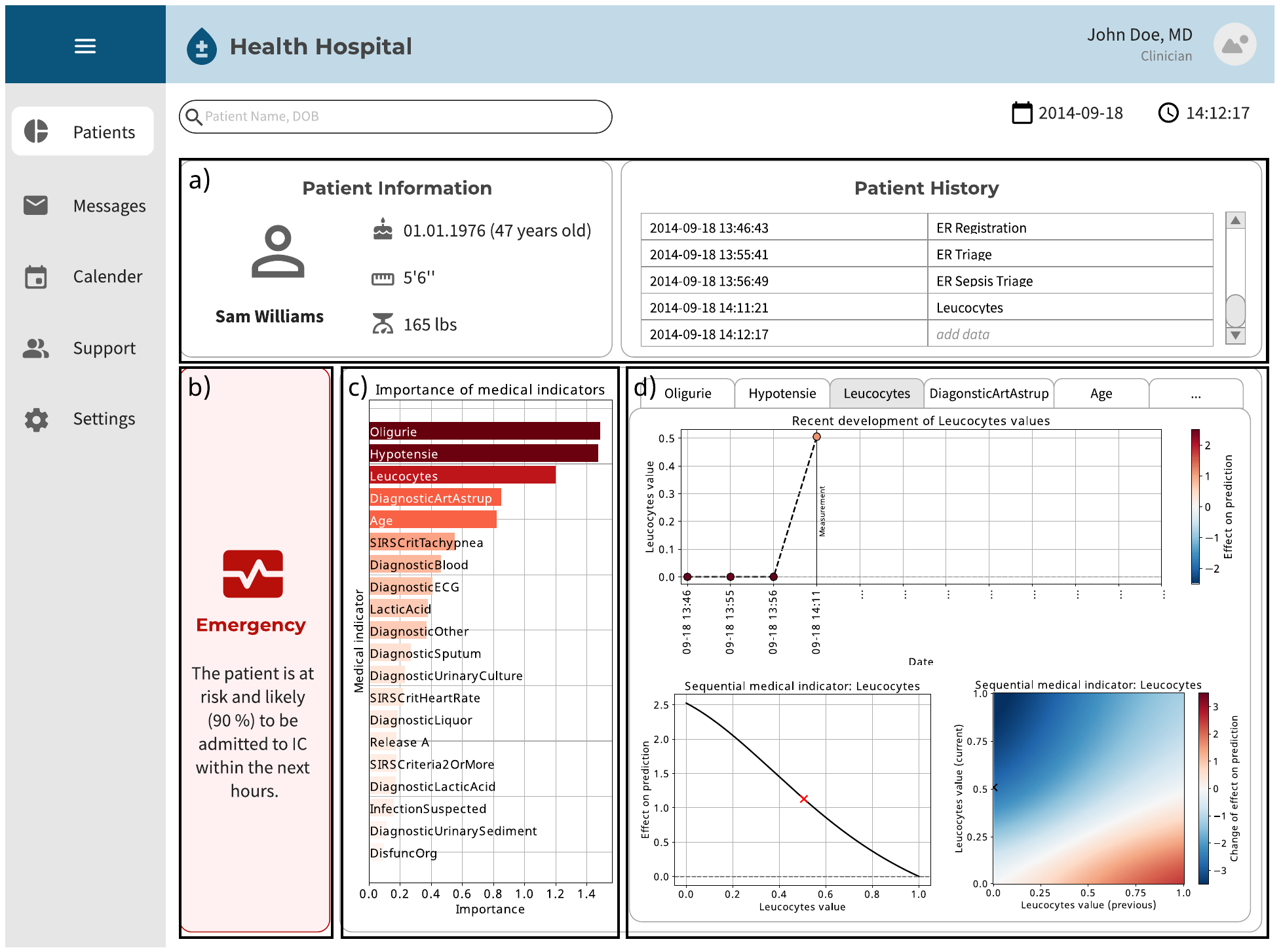}
    \caption{Medical dashboard with PatWay-Net's interpretation plots.}
    \label{fig:dasboard}
\end{figure}

PatWay-Net's interpretation plots are presented to medical decision-makers via a comprehensive dashboard (see~\Cref*{fig:dasboard}), which is structured into four parts, a) to d). 

Part a) provides general (static) information on a patient, such as age, height, or weight as well as their history during their current stay. For example, it shows when a patient has been admitted or when certain measurements have been taken. Part b) provides a short textual description of the urgency of the \ac{icu} admission depending on the model's prediction. Parts c) and d) comprise PatWay-Net's interpretation plots. In particular, part c) shows an overview of the most impactful medical indicators on the model prediction, and d) provides further interpretation details on selected static or sequential medical indicators.  
In what follows, we focus on parts c) and d) of the dashboard and demonstrate PatWay-Net's interpretation plots for the medical indicator importance as well as static and sequential medical indicators.

The interviews conducted with medical experts show that the dashboard is helpful as a support for decision-making. Moreover, all medical experts confirm the usefulness of the interpretation plots to understand at a glance what caused the prediction. The interviewees also positively assessed the visual plots and thought that such plots are the language that is spoken medically. All interviewees stated that they prefer simple plots because they usually have to act relatively quickly. Another outcome of the interviews is that interpretations in the form of PatWay-Net's dashboard would positively influence their trust in the predictions. Therefore, they think that it increases the acceptance of such predictions. Additional information on the interviews can be found in Appendix~\ref*{sec:eval_interviews}.

\subsubsection{Importance of medical indicators}

\begin{figure}[ht]

    \centering
    \includegraphics[width=0.29\columnwidth]{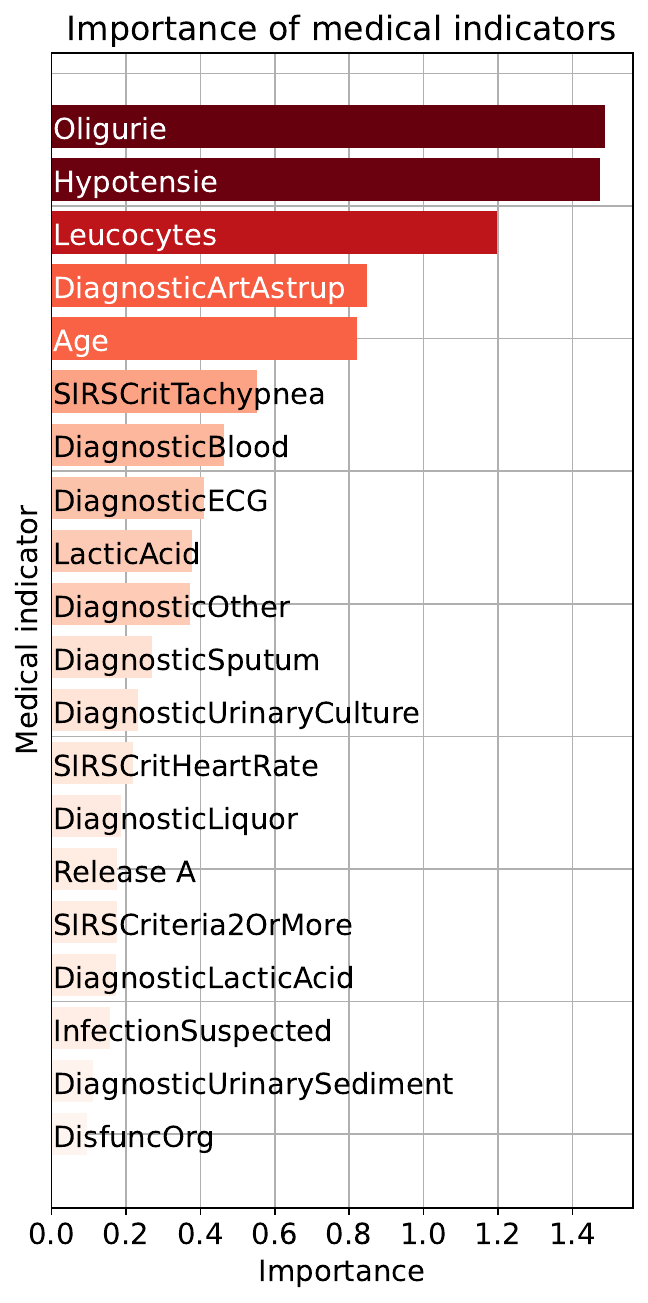}
    \caption{Importance for static and sequential medical indicators.}
    \label{fig:importance_med_indicators}
\end{figure}

\Cref*{fig:importance_med_indicators} shows the medical indicator importance plot, highlighting the 20 most impactful indicators in our model. These indicators have the greatest effects on the model output in forecasting the potential need for \ac{icu} admission.

The medical indicators \emph{Oligurie}, \emph{Hypotensie} (hypotension), and \emph{Leukocytes} emerge as the top contributors with the most substantial impact on the model prediction. 
The static indicator \emph{Oligurie}, representing decreased urine output, is an essential medical indicator in our model for predicting \ac{icu} admission as it is often associated with severe sepsis due to its connection with reduced kidney perfusion \citep[e.g.,][]{klein2018oliguria}. 
Likewise, \emph{Hypotensie}, or low blood pressure, is a critical static medical indicator in our model as it can represent a possible consequence of significant blood vessel dilation caused by systemic inflammation \citep[e.g.,][]{hotchkiss2016sepsis}.
The sequential indicator \emph{Leukocytes}, that is white blood cell count, also holds significant importance in our model for the prediction of \ac{icu} admission. Variations in leukocyte counts often signal the body's immune response to infections such as sepsis \citep[e.g.,][]{urrechaga2018role}. 
As our proposed \ac{ml} framework's unique capability is to include this sequential medical indicator in the analysis, it enables us to compare the importance of sequential indicators with the importance of the static indicators directly. 

\subsubsection{Static medical indicators}

\begin{figure}[ht]
    \centering
    \includegraphics[width=0.74\columnwidth]{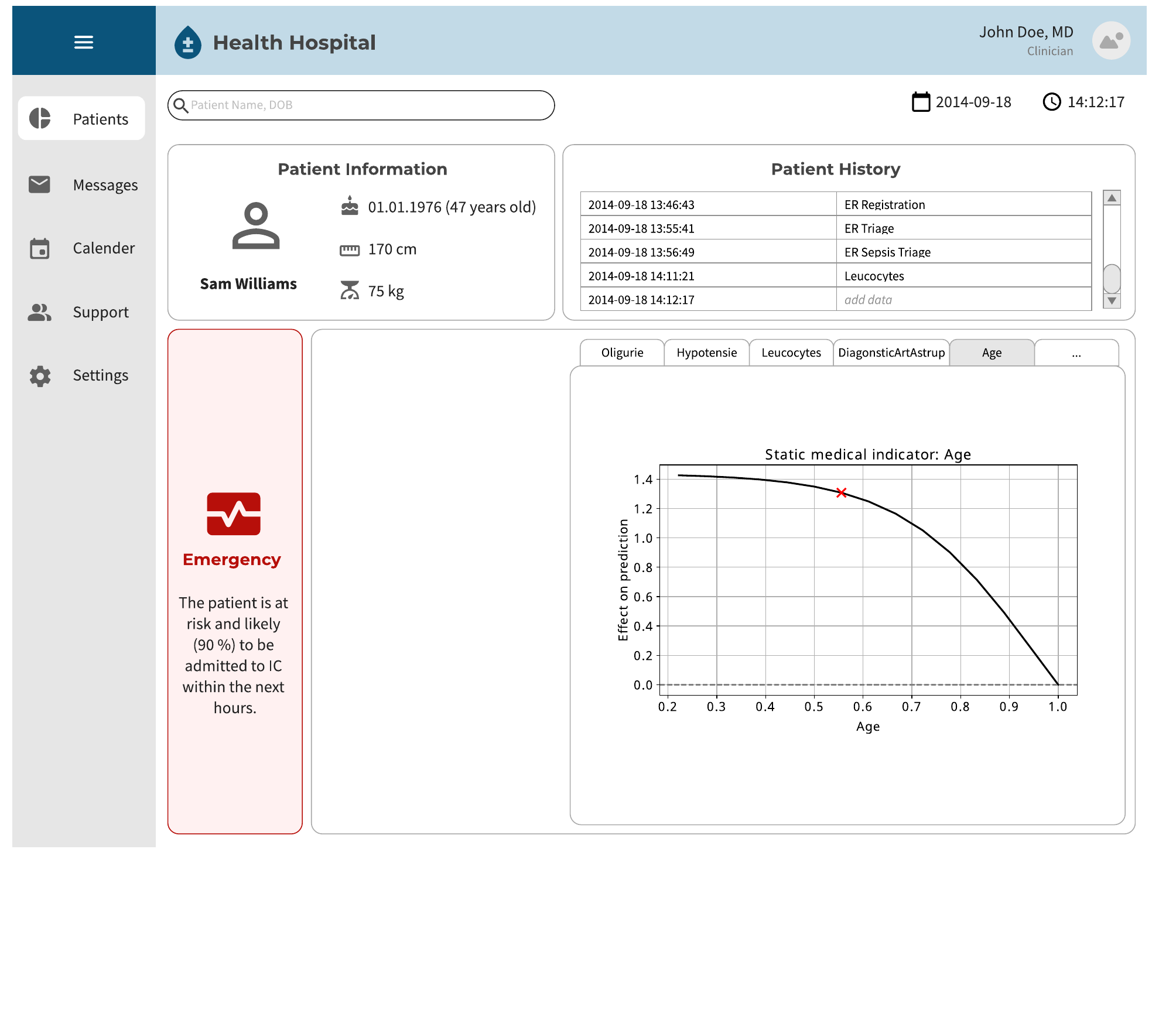}
    \caption{Interpretation plot for static medical indicator \emph{Age} in the dashboard.}
    \label{fig:Stat-Diag}
\end{figure}

\Cref*{fig:Stat-Diag} shows the interpretation plot for the static medical indicator \emph{Age} in the dashboard. Specifically, it shows PatWay-Net's shape function for this specific indicator, revealing a non-linear effect of \emph{Age} on \ac{icu} prediction. This demonstrates the flexibility of PatWay-Net's \ac{dnn} model in capturing arbitrary relationships between individual medical indicators and the prediction target, which, inspired by traditional \acp{gam} \citep[e.g.,][]{agarwal2021neural, kraus_interpretable_2023, yang2021gami, lou_accurate_2013, lou_intelligible_2012}, is more suitable for detecting and learning complex patterns in data than simple linear models.

Interestingly, the effect for the medical indicator \emph{Age} decreases as the value (i.e., the patient's age) increases. At first glance, this trend may appear counter-intuitive, considering that higher age is typically associated with more severe sepsis cases and a higher risk of adverse outcomes \citep[e.g.,][]{pera2015immunosenescence}. However, the patient population in our dataset is comparatively old, which might be a reason for this observed medical indicator shape. Additionally, there could be specific hospital protocols or clinical guidelines that apply to patients above a certain age, which could influence the patient's pathway and the eventual outcome. 

\subsubsection{Sequential medical indicators}

\begin{figure}[H]
    \centering
    \includegraphics[width=0.74\columnwidth]{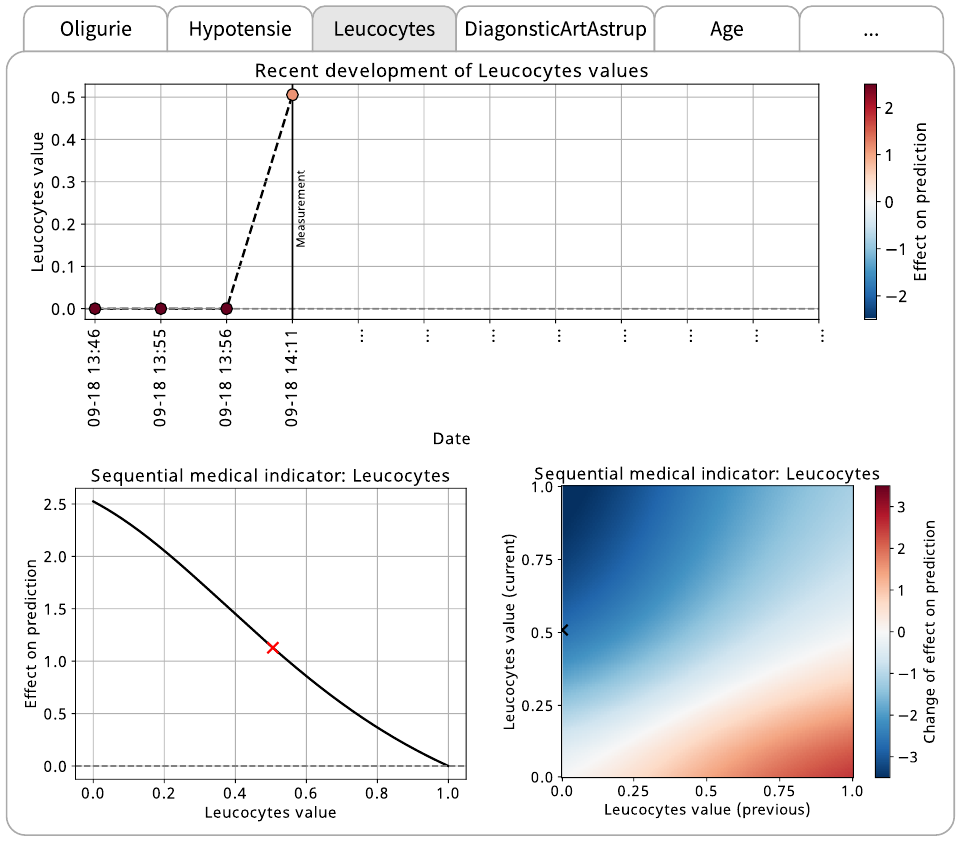}
    \caption{Interpretation plots for sequential medical indicator \emph{Leukocytes} in the dashboard.}
    \label{fig:Seq-Diag}
\end{figure}

\Cref*{fig:Seq-Diag} shows the interpretation plots for the sequential medical indicator \emph{Leukocytes} in the dashboard, depending on a given patient's previous pathway. The plots reveal the model's transparent decision logic regarding the development, shape, and transition of the medical indicator.  

The medical indicator shape plot (lower left in \Cref*{fig:Seq-Diag}) shows the shape function of the sequential medical indicator \emph{Leukocytes}. Again, we can see the flexibility of PatWay-Net's \ac{dnn} model in capturing non-linear relationships between the medical indicator and the prediction target. This time, however, the indicator represents a sequential feature captured in its natural form, which constitutes an innovative advancement over traditional interpretable models, such as \acp{gam} and decision trees.

\emph{Leukocytes} play a pivotal role in the body's immune response, and a considerable alteration in the leukocyte count is a common physiological response to sepsis \citep[e.g.,][]{urrechaga2018role}. 
In \Cref*{fig:Seq-Diag}, while the \emph{Leukocytes} value decreases, the effect on the prediction for \ac{icu} admission increases considerably. Thus, we can see a substantial alteration in the leukocyte count.   
Moreover, an elevated leukocyte count can be a typical indicator of an ongoing systemic inflammatory response to an infection, like sepsis \citep[e.g.,][]{riley2015evaluation}. 
However, a decrease in \emph{Leukocytes} can also occur in severe cases where the immune system is overwhelmed, indicating a worsening of the patient's condition \citep[e.g.,][]{belok2021evaluation}. In such an acute case, there exists a potential necessity for the patient to receive intensive care. 

The medical indicator transition plot (lower right in \Cref*{fig:Seq-Diag}) shows how the prediction changes from the previous to the current \emph{Leukocytes} value measurement. The figure illustrates that a decrease in the \emph{Leukocytes} value (from a previous value of 0.0 to a current value of 1.0) corresponds to an increased probability of the patient requiring \ac{icu} admission. 
This is consistent with clinical understanding, as a decrease in leukocytes often denotes a heightened vulnerability to developing an infection like sepsis, suggesting a more severe disease course that may require intensive care. Conversely, if there was a low \emph{Leukocytes} value at the previous time step that subsequently increases by the current time step to a normal value, prediction indicates a lower likelihood of the patient being transferred to the ICU. This could suggest that the patient's immune response is stabilizing, or the infection is being effectively controlled, thus reducing the necessity for intensive care.

The medical indicator development plot (upper plot in \Cref*{fig:Seq-Diag}) shows what effect the sequential medical indicator \emph{Leukocytes} has on the model prediction over time. Up to time step three (2014-09-18 13:46 - 2014-09-18 13:56), the effect of \emph{Leukocytes} is high since no measurement has been taken yet. From time step three to four (2014-09-18 13:56 - 2014-09-18 14:11), the effect on the prediction decreases, as a medium-high \emph{Leukocytes} value of 0.51 has been measured in this time period.

\section{Discussion and future work}
\label{sec:discussion}

\subsection{Implications for healthcare management and practice}
Our research has multiple implications for healthcare management and practice.
First, PatWay-Net supports a straightforward analysis of patient pathways using patients' historical event data. In this way, subjectivity is avoided, and manual effort can be reduced to a minimum in decision-making.
Likewise, our model provides high predictive performance in the context of patients with symptoms of sepsis without relying on explicit process knowledge. This allows flexibility for decision support applications in highly complex and dynamic healthcare environments.
In our case, experiments have shown that the predictive performance is superior to traditional approaches by combining patients' static features with sequential features in a \ac{dnn} architecture that remains fully interpretable. This is a great advantage because prediction tasks in the healthcare sector are usually dominated by linear and logistic regression models with underlying static features to ensure a high degree of transparency~\citep[e.g.,][]{lee_prediction_2020, Caruana.2015, Shipe2019, Bertoncelli.2020}.

At the same time, PatWay-Net can improve decision-making in both patient-specific and administrative decision contexts. 
For example, in a patient-specific decision context, a model interpretation for admission to \ac{icu} prediction may indicate an increase in a patient's probability of being transferred to the \ac{icu} after being treated with a certain medication. Based on this insight, medical experts have the chance to intervene and apply corrective treatments to prevent worse consequences.
In an administrative context, model interpretation could reveal shortcomings in the hospital's IT system. For instance, conflicting predictions between PatWay-Net and clinicians can be traced down to potentially missing patient information within an \ac{erp} system, allowing for optimization of hospital operations. 

Finally, PatWay-Net provides timely decision support.
From a technical point of view, PatWay-Net's inference time is similar to one of the shallow interpretable models as the underlying model of PatWay-Net represents a function mapping the data input to the prediction output. Compared to the inference time, the training time of PatWay-Net is considerably higher than the training time of the shallow interpretable models as PatWay-Net is a \ac{dnn} with a recurrent \ac{ilstm} cell. Further, the training time increases with each sequential feature as each sequential feature is passed through a single corridor in the \ac{ilstm} cell. However, for our purpose, the inference time is far more important than the training time as the models are created and trained before they are applied in an online mode where the models are used for providing effective decision support.    
On the other hand, PatWay-Net's interpretations can be immediately retrieved from the model itself. In doing so, it is considerably faster than applying a post-hoc explanation method such as \ac{shap} for reconstructing explanations for non-interpretable \ac{dnn} models.     

\subsection{Implications for research}
PatWay-Net combines two crucial streams of research.  
The first stream follows the idea that more complex models, such as \acp{dnn}, can naturally model specific structures of the underlying data and, thereby, increase predictive performance. Thus, PatWay-Net employs an \ac{ilstm} in its sequential module to model temporal structures of sequential data, and several \acp{mlp} in its static module to model non-linear structures of static data.  
The second stream follows the idea that explanations of complex models can never provide the same understanding as that of intrinsically interpretable models \citep{rudin2019stop, zschech_game_2022}. Consequently, approximated explanations of complex models should be avoided or used carefully. As a remedy, PatWay-Net remains fully interpretable and prevents uncontrolled interactions of static and sequential features by incorporating the main principle of \acp{gam} into its entire \ac{dnn} architecture. As such, the model also provides an extension to traditional \acp{gam}, which are unable to capture sequential data structures in their natural form \citep[e.g.,][]{agarwal2021neural, kraus_interpretable_2023, yang2021gami, lou_accurate_2013, lou_intelligible_2012}.

Within the realm of medical research, this work is aligned with emerging trends advocating for a shift from static, tabular data to multimodal data representation \citep{acosta2022multimodal}. Traditional approaches often simplify complex health data such as images and vital signs into aggregated statistics or explicit features, thereby losing important information and only capturing a snapshot of the patient's health. Our framework addresses this gap by accurately modeling health trajectories through both, sequential and static data. The architecture is not limited to merely processing patient pathway data but it can also be adapted to other temporal sequences commonly encountered in healthcare, such as data from wearable and ambient biosensors \citep{van2023glycaemic}. By facilitating a more rigorous representation of human data, we improve not only the predictive performance but also the clinical utility of \ac{ml} models in healthcare settings.

\subsection{Limitations and outlook}
As with any research, our work is not free of limitations. 
First, we focused in this paper on a use case of patients with symptoms of sepsis to demonstrate the benefits of PatWay-Net in cases where trust in the \ac{ml} system is crucial to allow for practical applications. However, the application of PatWay-Net is not limited to this use case but can also be used to predict process-related outcomes in other tasks or domains involving static and sequential features, as shown by the results of the additional use cases in Appendix~\ref*{app:furtherdata}. Here, we find mixed results, highlighting that full generalizability in other contexts requires further work. 

Second, the event log sample from the real-life data application was relatively small, and using this sample for training PatWay-Net showed a performance decrease from validation to test scores. This difference could be an indicator of model selection criterion overfitting \cite{cawley2010over}, which might affect PatWay-Net's generalizability to unseen data. However, to mitigate the effect of this overfitting type, we followed the suggestion from Cawley and Talbot~\cite{cawley2010over} and adopted solutions for the problem of overfitting to the training criterion. 
In particular, we tested model regularization, hyperparameter minimization, and early stopping~\cite{cawley2007preventing, cawley2010over, qi2004predictive}. Among these solutions, performing early stopping achieved the best predictive performance for our use case of patients with symptoms of sepsis. In addition, results that we obtained from a further use case on loan applications (see Appendix~\ref*{app:furtherdata}) confirm that this type of overfitting is likely to be less present when the event log size is larger.
However, despite these overfitting concerns, PatWay-Net achieved relatively high predictive performance, and being a neural network, it can be expected that its predictive performance will further improve when trained with more data~\cite{lecun.2015}.

Third, PatWay-Net's sequential medical indicator plots provide interpretations that are tied to a patient's individual pathway. This limitation is necessary because the predictive effect of a sequential medical indicator within our \ac{ilstm} cell is determined by the patient-specific trajectory over previous time steps. As a result, varying historical trajectories can lead to different outcomes, which may also affect the results of the interpretation plots. However, at this point, it is not practical for clinical decision support to include global interpretation plots for all conceivable trajectory variants across all patients in a single dashboard. Therefore, we decided to focus on developing a patient-specific dashboard with all relevant information to support clinicians in an easily accessible way. Nonetheless, future research should address this limitation to identify new ways of how feature effects of multiple sequences over several time steps can be visualized in a comprehensive, yet fully understandable manner. This may require new visualization techniques (e.g., interactive filter mechanisms) or additional abstraction layers (e.g., clustering of patient trajectories leading to similar outcomes and interpretation plots), which offer promising directions for future work.

Fourth, PatWay-Net's mechanism to automatically detect and integrate interactions covers pairwise interactions among sequential features. For the use case addressed in this paper, we can show that the predictive performance of PatWay-Net with this mechanism is close to the predictive performance of an unrestricted \ac{lstm} cell (see Appendix~\ref*{app:use_case}). Nevertheless, we assume that other types of interactions (e.g., more complex interactions between sequential features or interactions between sequential and static features) are more present in other use cases. The results obtained for a further use case on hospital billings (see Appendix~\ref*{app:furtherdata}) give the first indication for this assumption and therefore provide an entry point for future research.     

Fifth, the proposed version of PatWay-Net does not currently consider a mechanism for selecting relevant features. This may become relevant when dealing with a large collection of features in other real-world applications, where the full set of features may lead to impractically large computational costs and a higher risk of overfitting. Future research could follow up on this point to investigate which feature selection methods are appropriate for combining static and sequential features. Nevertheless, PatWay-Net already provides some guidance for selecting the most important features (or medical indicators) through its medical indicator importance plot, thus facilitating clinicians or hospital management when dealing with a large collection of medical indicators. 

Sixth, as with any \ac{ml} model, PatWay-Net's results are only as good as the data it consumes. That is, PatWay-Net is not only a reflection of possibly biased decisions made in the past but also of any data quality issues embedded in the data set. For example, in our use case of patients with symptoms of sepsis, some interpretation plots showed counter-intuitive relationships between medical indicators and the prediction target that may not be reflected in the medical literature. These findings underscore the need for rigorous data management in hospital operations to enable analytics tools like PatWay-Net to enhance decision-making substantially. Similarly, we want to emphasize that the learned feature effects should not be interpreted causally, as they are still based on correlations. Thus, it is not possible to say with certainty why some of the effects shown in the interpretation plots are present. This could be due to correlations with other (unmeasured) features, or other underlying phenomena. However, despite these limitations, PatWay-Net still offers a fully transparent model that can be used to allow clinicians to compare the model results with their domain knowledge to iteratively debug and improve the model, identify underlying data quality issues, or initiate further investigations for the identification of causal relationships.

Finally, our current approach pertains to the creation of a clinical dashboard that relies solely on the provided interpretation plots generated from the shape functions of PatWay-Net. While these plots offer exact insights into the model's decision logic, they may still lack the level of context and intuitiveness required for effective clinical application. In the next steps, we intend to address this limitation by harnessing the capabilities of large language models \citep{slack_explaining_2023, feuerriegel_generative_2023}. By incorporating a large language model in an adaptive dialogue system, we aim to provide more intuitive and contextually relevant explanations for clinical professionals when presenting the interpretation plots. This enhancement will not only make the model's outputs more accessible but also foster improved communication between the model and the healthcare practitioners, thereby enhancing the model's utility in real-world clinical settings.

\clearpage
\newpage

\section*{Data and code availability}

\textbf{Data availability} Only a public data set was used, as cited in the text.
\\
\noindent
\textbf{Code availability} The code is available under the link provided in the text.

\section*{Declarations}

\textbf{Conflict of interests} The authors declare that they have no conflict of interest.\\
\noindent
\textbf{Funding} Mathias Kraus and Patrick Zschech acknowledge funding from the Federal Ministry of Education and Research (BMBF) on ``White-Box-AI” (Grant 01IS22080). Patrick Zschech acknowledges funding from the Federal Ministry of Education and Research (BMBF) on ``AddIChron” (Grant 16SV8995). Martin Matzner acknowledges funding from the German Reseach Foundation (DFG) on ``CoPPA" (Grant 456415646).

\newpage
\begin{appendices}

\section{Further details and experiments on the main use case}
\label{app:use_case}

In this appendix, we provide further details on the use case we use to evaluate the clinical utility of PatWay-Net, our proposed \ac{ml} framework for interpretable predictions in patient pathways.

In what follows, we provide details on the preprocessing of the data set~(Appendix~\ref*{app:preprocessing}), before we present a heuristic for automatic interaction detection (Appendix~\ref*{app:automatic_search_interactions}). After that, we provide details on model tuning, model evaluation, and model selection (Appendix~\ref*{app:hyperparameter_tuning}), and present further results on statistical tests (Appendix~\ref*{app:stattest}), predictive performance~(Appendix~\ref*{app:multiple_interactions}, \ref*{app:predprefsize}, and \ref*{app:predpreftime}), interpretation quality~(Appendix~\ref*{app:inter-ex-compare}), and runtime performance~(Appendix~\ref*{sec:runtime}).   

\subsection{Preprocessing of the data set}
\label{app:preprocessing}
We remove outliers in our data set by only considering completed patient pathways that are longer than two but shorter or equal to 50 patient events. We also remove patient pathways that do not start with activity \emph{ER registration} because we assume this activity to be the central entry point into the patient pathway. As a result, the event log contains 724 patient pathways with 675 different variants over a period of 1.5 years.

Our real-life data set comprises binary, categorical, or continuous medical indicators. The values of a binary medical indicator are mapped to 0 or 1, categorical values are onehot-encoded, and continuous values are scaled into the range $[0, 1]$. Patient activities, such as \emph{\ac{icu} admission} or \emph{Measure blood pressure}, are either encoded by standard onehot encoding or by a custom encoding. In standard one-hot encoding, the patient activity $a$ is encoded as a vector containing only zeros, except for a single position that corresponds to $a$, which is set to 1. In our custom encoding, we explicitly model the relationship between activities and their existing continuous medical indicators in the data. In detail, if an activity can be described by a continuous value, we set the corresponding position in the vector to this continuous value. Further, to provide more meaningful interpretations for sequential medical indicators, we also keep this continuous value for subsequent patient activities, as long the value does not change. 

We extract all prefixes from $\mathbf{X}_{static} \times \mathsf{X}_{seq}$; that is, we extract all subsequences of the sequential data, denote those as $\mathsf{X}_{seq}^{sub}$, and retain the static data as is, to predict at each time step. This step increases the number of training samples, which enables us to evaluate the \ac{ml} models on how early they can already make accurate patient pathway predictions in the future. 

For each patient pathway prefix, the target labels $\mathbf{y}$ are created as follows: Given a prefix \mbox{$(\mathbf{X}_{static}^{(i)}, \, \mathsf{X}_{seq}^{(i)}[:t^*])$} and the patient activity of interest (e.g., Admission to \ac{icu}), we check the activities of patient pathway $i$. Then, if the activity of interest appears in the patient pathway’s activities, we set the target label to 1 and else to 0. 
We discard all prefixes and corresponding labels where the activity of interest is part of the sequential data. This is important to avoid data leakage problems in patient pathway predictions.

\subsection{Automatic search for interactions}
\label{app:automatic_search_interactions}
Given all subsequences of the sequential data $\mathsf{X}_{seq}^{sub}$, PatWay-Net's interaction detection iterates 100 times to identify the most relevant pairwise feature interactions in these data (see \Cref*{alg:model_inter}). Per iteration, a feature pair $(j,k)$ is randomly determined from sequential features $\mathcal{D}_{seq}$, and the sequential data for the features $j$ and $k$ are retrieved and reshaped. Then, the sequential data and label data are split into an 80\% training and 20\% test set, an XGBoost~\citep{chen2016xgboost} model $f_{XGB}$ with standard parameters is trained based on this data, and the trained model is applied to the test set to calculate an $AUC_{ROC}$ value. Subsequently, the current interaction is added together with the respective $AUC_{ROC}$ value to $\mathbf{r}$, from which the best interactions are selected. In addition, the current interaction is added to $\mathcal{K}$ so that the same interaction cannot be used again in future iterations of this procedure. After performing all iterations, the interactions with the highest $AUC_{ROC}$ values are first selected based on $\mathbf{r}$ and $k^{*}$ (number of best interactions) and then transferred to the \ac{ilstm} layer, in which they are considered as additional sequential features.

\begin{algorithm}[ht]
\small
\DontPrintSemicolon
\SetAlgoLined
\setcounter{AlgoLine}{0}
\LinesNumbered
\SetKwInput{Result}{Result}
\SetKwInOut{Given}{Given}
\Given{$\mathsf{X}_{seq}^{sub}, \mathbf{y}, k^{*}, \mathcal{D}_{seq}, f_{XGBoost}.$}
\For{$i \gets1$ \KwTo 100}{
    $(j,k) \gets \text{feature pair}(\mathcal{D}_{seq}).$
    
\If{$j \neq k \wedge (j,k) \not\in \mathcal{K} $}{
    
    $\mathsf{X}_{seq}^{sub} \gets (\mathsf{X}_{seq}^{sub}[j],\,\mathsf{X}_{seq}^{sub}[k])$.
    
    $\mathbf{X}^{sub}_{seq} \gets \text{reshape}(\mathsf{X}_{seq}^{sub})$.
    
$\mathbf{X}^{train}_{seq}, \,\mathbf{y}^{train} \gets \text{split}(\mathbf{X}_{seq}^{sub}, \,\mathbf{y}).$

$\mathbf{X}^{test}_{seq}, \,\mathbf{y}^{test} \gets \text{split}(\mathbf{X}_{seq}^{sub}, \,\mathbf{y}).$

${f}_{XGBoost}^{*} \leftarrow \text{train}(f_{XGBoost}, \, \mathbf{X}^{train}_{seq}, \mathbf{y}^{train})$.

$perf_{i} \gets \text{test}({f}_{XGBoost}^{*}, \, \mathbf{X}^{test}_{seq}, \mathbf{y}^{test})$.

$\mathbf{r} \gets \mathbf{r} + (perf_i, \, (j,k)).$

$\mathcal{K} \gets \mathcal{K} \cup \{(j,k)\}.$

}
}
\Return{$\text{top-k-inter}(\mathbf{r}, k^{*}).$}
\caption{PatWay-Net's interaction detection.}
\label{alg:model_inter}
\end{algorithm}

\subsection{Model tuning, model evaluation, and model selection}
\label{app:hyperparameter_tuning}
\Cref*{tab:hpo} reports the tuning parameters used in our grid search. 
We performed initial experiments for all models to find appropriate value ranges for the hyperparameters.  
For the PatWay-Net models, the hyperparameters \emph{hidden size per sequential feature} and \emph{hidden size per static feature} define the used vector space for each sequential and static feature, respectively. For example, if the hidden size per static feature is set to 4, each model's \ac{mlp} has a hidden layer with 4 neurons. In contrast to the PatWay-Net models, the baseline \ac{lstm} model uses a traditional unrestricted \ac{lstm} cell instead of the proposed, restricted iLSTM cell, and therefore, the hidden size per sequential feature defines the used vector space for all sequential features. 
Further, we set the maximum number of neurons for the \ac{lstm} model with the unrestricted \ac{lstm} cell to 8 and for the PatWay-Net models with the restricted \ac{ilstm} cell to 128 as experiments in the use case showed that more vector space is required for the unrestricted \ac{lstm} to identify arbitrary dependencies between all sequential features and less vector space is sufficient for the restricted \ac{lstm} to compute each sequential feature effectively and efficiently.      
For the shallow \ac{ml} models, we set the value ranges of the hyperparameters such that the resulting models are still interpretable. For instance, we bounded the depth of the decision tree or the number of neighbors in the $K$-nearest neighbor model.

\begin{table}[ht]
	\begin{center}
\caption{Hyperparameters used in grid search. Best values over the five folds are marked in bold.}
\label{tab:hpo}
\begin{tabular}{@{}lll@{}}
\toprule
\ac{ml} approach & Hyperparameter & Hyperparameter range \\ \midrule
PatWay-Net & Hidden size per sequential feature & 4, \textbf{8} \\
(with interaction) & Hidden size per static feature & 4, \textbf{8}\\
& Learning rate & 0.001, \textbf{0.01} \\
& Batch size & \textbf{32}, 128 \\
PatWay-Net & Hidden size per sequential feature & \textbf{4}, 8 \\
(without interaction) & Hidden size per static feature & \textbf{4}, 8\\
& Learning rate & \textbf{0.001}, 0.01 \\
& Batch size & \textbf{32}, 128 \\
\ac{lstm} network & Hidden size sequential features & 4, 32, \textbf{128}\\
(with static module) & Hidden size per static feature & \textbf{4}, 8\\
 & Learning rate & \textbf{0.001}, 0.01 \\
 & Batch size & \textbf{32}, 128 \\
Decision tree & Max. depth & 2, 3, \textbf{4}\\
Logistic regression & Regularization strength & $10^{-3}, {10^{-2}}, \dots,\mathbf{10^{0}} \dots,10^{+3}$\\
$K$-nearest neighbor & Number of neighbors & 3, 5, \textbf{10} \\
Na\"ive Bayes & Variance smoothing& $10^{-9}, 10^{-8}, \dots,\mathbf{10^{-4}},\dots,1$\\
XGBoost & Max. depth & \textbf{2}, 6, 12 \\
 & Learning rate & 0.3, 0.1, \textbf{0.03}\\
 Random forest & Max. depth & 2, 6, \textbf{12} \\
  & Number of estimators & 100, 200, \textbf{400} \\
  & Max. leaf nodes & 2, 6, \textbf{12} \\

\bottomrule
\end{tabular}%
\end{center}
\end{table}

The procedure for model evaluation is summarized in \Cref*{alg:model_evaluation}. Given the static data $\mathbf{X}_{static}$, sequential data $\mathsf{X}_{seq}$, and target outcome $\mathbf{y}$, we start our evaluation by performing a five-fold stratified cross-validation with random shuffling on patient pathway level. That is, training and validation of each run are performed based on an entire pathway, without randomly shuffling sequentially ordered events, to avoid any temporal data leakage that could result from erroneously using future events as part of the evaluation of past events.
After retrieving the training and test set in each fold, we split the training set into a sub-training set (train*) and a validation set. Subsequently, we select the best model through a grid search using the sub-training and validation set and apply the best model to the test set to compute the test performance. We perform this procedure five times ($k = 5$). In total, we repeat the entire model evaluation procedure five times, each with a different seed, and calculate the average performance and standard deviation over the performance values of these executions.

\begin{algorithm}[!ht]
\small
\DontPrintSemicolon
\SetAlgoLined
\setcounter{AlgoLine}{0}
\LinesNumbered
\SetKwInput{Result}{Result}
\SetKwInOut{Given}{Given}
\Given{$\mathbf{X}_{static}$, $\mathsf{X}_{seq}$, $\mathbf{y}, k, f$}

$(tr_{1}, te_{1}), \dots ,(tr_{k}, te_{k}) \gets \text{split-k-stratified}(\mathbf{X}_{static}, \mathbf{y}).$

\For{$i \gets1$ \KwTo $k$}{

$\mathbf{X}^{train}_{static}, \mathsf{X}^{train}_{seq}, \mathbf{y}^{train} \gets \text{retrieve}(\mathbf{X}_{static}, \mathsf{X}_{seq}, \mathbf{y}, tr_{i}).$

$\mathbf{X}^{test}_{static}, \mathsf{X}^{test}_{seq}, \mathbf{y}^{test} \gets \text{retrieve}(\mathbf{X}_{static}, \mathsf{X}_{seq}, \mathbf{y}, te_{i}).$

$\mathbf{X}^{train*}_{static}, \mathsf{X}^{train*}_{seq}, \mathbf{y}^{train*} \gets \text{split}(\mathbf{X}^{train}_{static}, \mathsf{X}^{train}_{seq}, \mathbf{y}^{train}).$

$\mathbf{X}^{val}_{static}, \mathsf{X}^{val}_{seq}, \mathbf{y}^{val} \gets \text{split}(\mathbf{X}^{train}_{static}, \mathsf{X}^{train}_{seq}, \mathbf{y}^{train}).$

 ${f}_{best} \leftarrow$ \text{grid-search}$(f, \mathbf{X}^{train*}_{static}, \mathsf{X}^{train*}_{seq}, \mathbf{y}^{train*}, \mathbf{X}^{val}_{static}, \mathsf{X}^{val}_{seq}, \mathbf{y}^{val})$
 
 $perf_{i} \leftarrow \text{test}({f}_{best}, \mathbf{X}^{test}_{static}, \mathsf{X}^{test}_{seq}, \mathbf{y}^{test}).$
 
}
\Return{$\frac{1}{k} \sum^{k}_{i=1} perf_{i}.$}
\caption{Model evaluation.}
\label{alg:model_evaluation}
\end{algorithm}

To measure the performance of the \ac{ml} models on the validation and the test set, we calculate the $AUC_{ROC}$~\citep{davis.2006} (primary measure) and the weighted F1-score (secondary measure). $AUC_{ROC}$ determines how well a classifier can distinguish between classes~\citep{mcclish1989analyzing}, and remains unbiased when dealing with highly imbalanced class distributions~\citep{bradley.1997}. F1-score is the harmonic mean of precision and recall~\citep{davis.2006}. 
Moreover, we select the model with the highest test $AUC_{ROC}$ from the model evaluation procedure to retrieve the interpretation. 

For model selection, we perform a grid search, as formalized in \Cref*{alg:model_selection}.
\begin{algorithm}[ht]
\small
\DontPrintSemicolon
\SetAlgoLined
\setcounter{AlgoLine}{0}
\LinesNumbered
\SetKwInput{Result}{Result}
\SetKwInOut{Given}{Given}
\Given{$\mathbf{X}^{train}_{static}, \mathsf{X}^{train}_{seq}, \mathbf{y}^{train}, \mathbf{X}^{val}_{static}, \mathsf{X}^{val}_{seq}, \mathbf{y}^{val}, f, \mathcal{P}$.}

\For{$(p_{1}, \dots, p_{l}) \in \mathcal{P}_{1} \times, \dots, \times \mathcal{P}_{l}$}{

${f}_{(p_{1}, \dots, p_{l})} \leftarrow \text{train}(f, (p_{1}, \dots, p_{l}), \mathbf{X}^{train}_{static}, \mathsf{X}^{train}_{seq}, \mathbf{y}^{train}).$

$AUC_{ROC,(p_{1}, \dots, p_{l})} \leftarrow \text{validate}({f}_{(p_{1}, \dots, p_{l})}, \mathbf{X}^{val}_{static}, \mathsf{X}^{val}_{seq}, \mathbf{y}^{val}).$

\If{$AUC_{ROC,(p_{1}, \dots, p_{l})} > AUC_{ROC,best}$}{

${f}_{best} \leftarrow {f}_{(p_{1}, \dots, p_{l})}.$

$AUC_{ROC,best} \leftarrow AUC_{ROC,(p_{1}, \dots, p_{l})}.$
}
}
\Return{$f_{best}$}
\caption{Model selection (grid-search).}
\label{alg:model_selection}
\end{algorithm}
For each hyperparameter constellation $(p_{1}, \dots, p_{l}) \in \mathcal{P}$, the model $f$ is trained and validated. During training, we perform early stopping based on the validation set after 10 epochs to avoid overfitting. After training, we apply the model to the validation set and calculate the $AUC_{ROC}$. The $AUC_{ROC}$ is used as our selection criterion for the grid search, and the model with the highest $AUC_{ROC}$ value on the validation set is selected for model testing.

\subsection{Statistical tests}
\label{app:stattest}
We perform statistical tests for the interpretable \ac{ml} approaches used in our real-life data application. In particular, for our main metric (test $AUC_{ROC}$), we conduct a Friedman test showing significant differences among the results (statistic = 56.45143, p-value = 6.56037e-11). As a post-hoc test, we conduct a Wilcoxon signed-rank test with Holm p-value adjustment \citep{demsar2006}. \Cref*{tab:stat_test} shows the pairwise p-values. 
PatWay-Net with one interaction outperforms the decision tree ($p=.001503$), logistic regression ($p=.000593$), and $K$-nearest neighbor ($p=.000001$) models significantly with $\alpha = 1\%$ and the setting with one interaction significantly outperforms the na\"ive Bayes ($p=.044226$) models with $\alpha = 10\%$. PatWay-Net's setting without interaction also outperforms the decision tree ($p=.000637$), logistic regression ($p=.002009$), and $K$-nearest neighbor ($p=.000001$) models significantly with $\alpha = 1\%$.

\begin{table}[ht]
\centering
\caption{Overview of pairwise p-values for test $AUC_{ROC}$ for Wilcoxon signed-rank test.}
\label{tab:stat_test}
\resizebox{\textwidth}{!}{
\begin{tabular}{@{}rrrrrrr@{}}
\toprule
\multirow{2}{*}{p-values for Test $AUC_{ROC}$}   &PatWay-Net  & {PatWay-Net} & Decision       & $K$-nearest     & Na\"ive       & Logistic          \\ 

    & (with int.) & (without int.) &  tree      & neighbor     &   Bayes     &   regression     \\
\midrule
PatWay-Net (with int.) &                                   & {.379944}                       & {.001503} & {.000001} & {.044226} & {.000593} \\
PatWay-Net (without int.) & {.379944}  &  & {.000637} & {.000001} & {.118250} & {.002009} \\
Decision tree & .001503                          & .000637                       &          & .001461 & .340556 & .915693 \\
$K$-nearest neighbor & .000001   & .000001   & .001461 &          & .000160 & .000383 \\
Na\"ive Bayes    & .044226      & .118250    & .340556 & .000160 &          & .915693 \\
Logistic regression                                & .000593                          & .002009                       & .915693 & .000383 & .915693 &   \\

\bottomrule
\end{tabular}
}
\end{table}

\subsection{Predictive performance for controlled vs. uncontrolled interactions}
\label{app:multiple_interactions}
\Cref*{tab:multiple_interaction_results} describes the results for PatWay-Net, in which the number of interactions varies. To provide a fair comparison, the experiments for PatWay-Net with two and three interactions are performed in the same way as described in Appendix~\ref*{app:hyperparameter_tuning}.
Overall, we observe robust results with only marginal differences. We find that in our real-life use case, a single sequential interaction leads to the highest $AUC_{ROC}$ and F1-score on the test~set.

\begin{table}[ht]
\caption{Comparison of PatWay-Net with different interactions and the \ac{lstm} network (with static module).}
\label{tab:multiple_interaction_results}
\centering
\resizebox{\textwidth}{!}{
\begin{tabular}{@{}lrrlrr@{}}
\toprule
\multirow{2}{*}{\begin{tabular}{l}\ac{ml} approach\end{tabular}}  &\multicolumn{2}{c}{F1-score (weighted)}                   &  & \multicolumn{2}{c}{$AUC_{ROC}$}                               \\ \cmidrule(lr){2-3} \cmidrule(l){5-6}

                        &   \multicolumn{1}{c}{Validation} & \multicolumn{1}{c}{Test} &  & \multicolumn{1}{c}{Validation} & \multicolumn{1}{c}{Test} \\ \midrule
\multicolumn{6}{l}{{\textsc{Controlled Interactions in our approach}}}\\
\begin{tabular}{l}PatWay-Net (one interaction)\end{tabular}         & 0.886 ($\pm$.016)       & 0.896 ($\pm$.016) &  & 0.820 ($\pm$.028)       & 0.734 ($\pm$.058) \\
\begin{tabular}{l}PatWay-Net (two interactions)\end{tabular}           & 0.886 ($\pm$.018)        &  0.894 ($\pm$.013) &  & 0.821 ($\pm$.035)       & 0.720 ($\pm$.045) \\
\begin{tabular}{l}PatWay-Net (three interactions)\end{tabular}        & 0.884 ($\pm$.017)       & 0.892 ($\pm$.015) &  & 0.817 ($\pm$.035)     & 0.718 ($\pm$.042) \\
\multicolumn{6}{l}{{\textsc{Uncontrolled Interactions in Non-Interpretable Machine Learning}}}\\

\begin{tabular}{l}\ac{lstm} network\\(with static module) \end{tabular}               &    0.890 ($\pm$.018)                            &   0.898 ($\pm$.014)                       &  &        0.840 ($\pm$.028)                        &   0.757 ($\pm$.049)                       \\

\bottomrule
\end{tabular}
}

\end{table}

\begin{figure}[ht] 
    \centering
    \includegraphics[height=6cm]{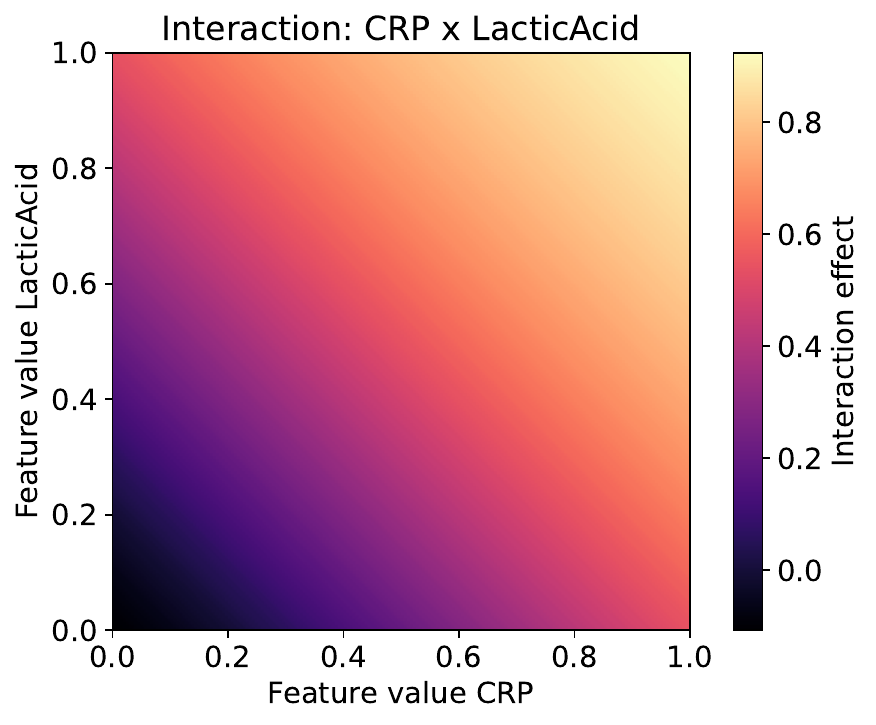}
    \caption{Feature interaction between the indicators \emph{\ac{crp}} and \emph{LacticAcid}.}
    \label{fig:interaction}
\end{figure}

Further, \Cref*{fig:interaction} illustrates the most relevant pairwise sequential medical indicator interaction in terms of the predictive performance of PatWay-Net (with one interaction). The figure demonstrates the interaction between the indicators \emph{\ac{crp}} ($x$-axis) and \emph{LacticAcid} ($y$-axis). If both indicator values are low, the interaction effect is low (black-colored region in the lower left corner). However, the interaction effect becomes stronger with increasing values of both indicators. So, if the values for both indicators are close to 1, the interaction effect is high (yellow-colored region in the upper right corner).

\subsection{Predictive performance for different training sample sizes}
\label{app:predprefsize}
\begin{figure}[ht]
    \centering
    \includegraphics[width=0.6\textwidth]{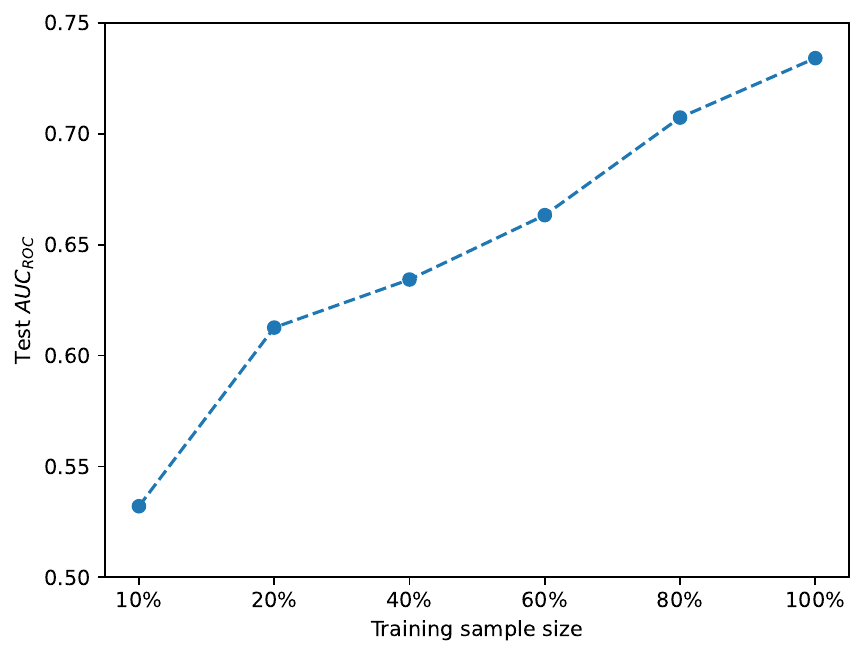}
    \caption{Predictive performance for different training sample sizes.}
    \label{fig:pred_perf_sample}
\end{figure}

\Cref*{fig:pred_perf_sample} shows the test $AUC_{ROC}$ scores of different training sample sizes for PatWay-Net with one interaction, the best-performing variant. More specifically, we create training samples with 10\%, 20\%, 40\%, 60\%, 80\%, and 100\% of the instances of the complete training set. To avoid data leakage issues, we create the training set samples based on complete patient pathways and split each of them into a train and validation set before the prefixes are created from the complete patient pathways. For each sample, we perform a five-fold cross-validation, as described in Appendix~\ref*{app:hyperparameter_tuning}, but with default hyperparameters. 

The figure shows that the test $AUC_{ROC}$ increases steadily with more training instances. Given that, we conclude that the event log size has an impact on the predictive performance and more training instances lead to a higher predictive performance. 
Further, as PatWay-Net outperforms all interpretable shallow \ac{ml} baseline models in our comparison, we conclude that the use of the complete training set is appropriate for PatWay-Net to create predictions that outperform the predictions of interpretable \ac{ml} baselines regarding $AUC_{ROC}$. 
Finally, as PatWay-Net belongs to the family of \acp{dnn} \citep{lecun.2015}, we assume that it achieves an even higher predictive performance and maybe a greater difference in predictive performance compared to the interpretable \ac{ml} baselines when more data are used for model training.

\subsection{Predictive performance over time}
\label{app:predpreftime}
\begin{figure}[ht]
    \centering
    \includegraphics[width=0.9\textwidth]{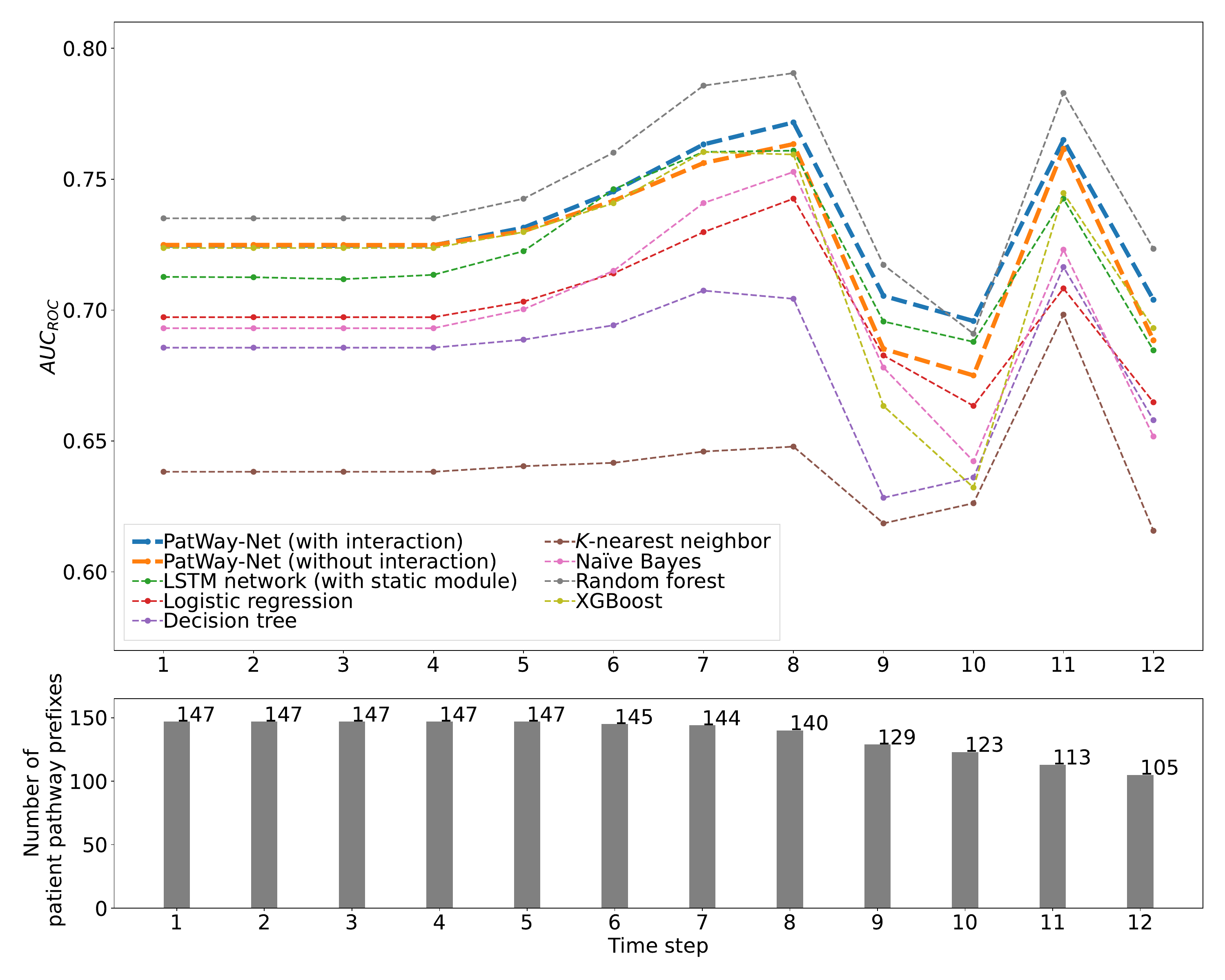}
    \caption{Predictive performance over time.}
    \label{fig:pred_perf_over_time}
\end{figure}

\Cref*{fig:pred_perf_over_time} shows the test $AUC_{ROC}$ scores of the first 12 time steps of patient pathways from the use case for PatWay-Net (without and with interaction) and the baselines. At each time step, prefixes of patient pathways of the corresponding size are considered. For calculating the $AUC_{ROC}$ scores per time step, we tune, evaluate, and select the models as described in \Cref*{app:hyperparameter_tuning}.         

The figure shows that PatWay-Net (with and without interaction) is already able to create predictions in the first steps of patient pathways, which are considerably more accurate in terms of $AUC_{ROC}$ than the predictions of the interpretable \ac{ml} baselines. Only the not-interpretable \ac{ml} approach random forest outperforms PatWay-Net (with and without interaction) in nearly all of the considered time steps. By considering only longer sequences (longer than 8), a significant number of patient paths are filtered out from this experiment, and thus the performance varies for all ML models. 

Overall, this experiment empirically shows that the size of the given event log is appropriate for PatWay-Net to create timely predictions.

\subsection{Intrinsic interpretability vs. post-hoc explainability}
\label{app:inter-ex-compare}
\begin{figure}[ht]
    \centering
    \begin{subfigure}{
        \includegraphics[height=3.7cm]{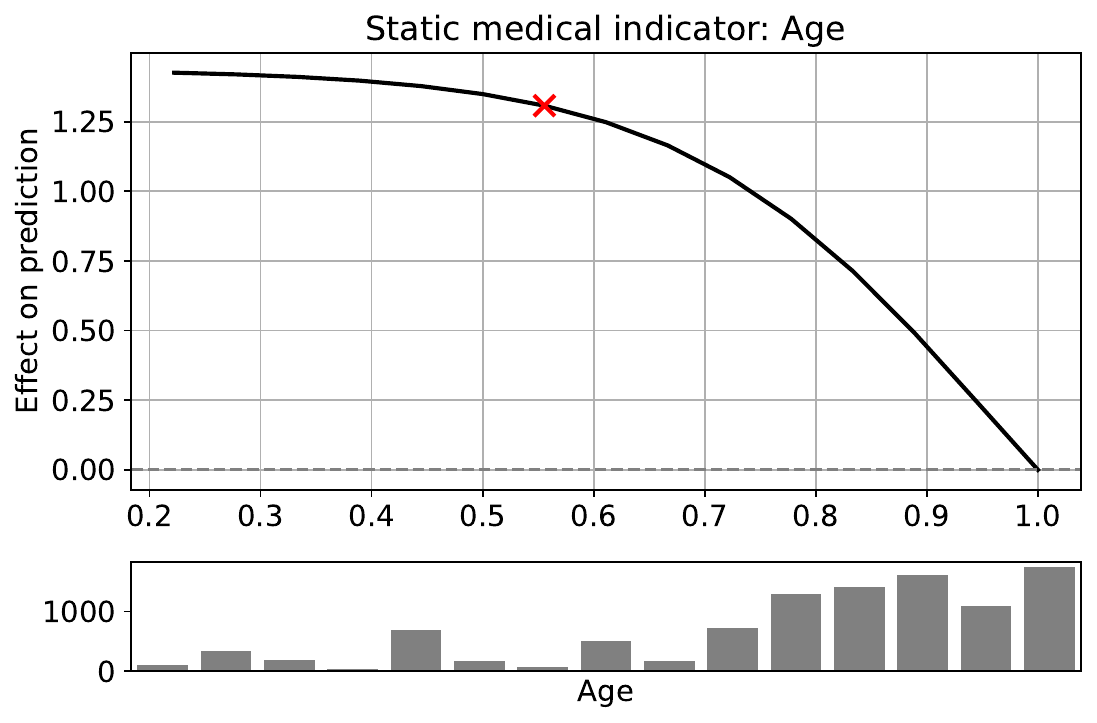}}
    \end{subfigure}%
    \begin{subfigure}{
        \includegraphics[height=3.7cm]{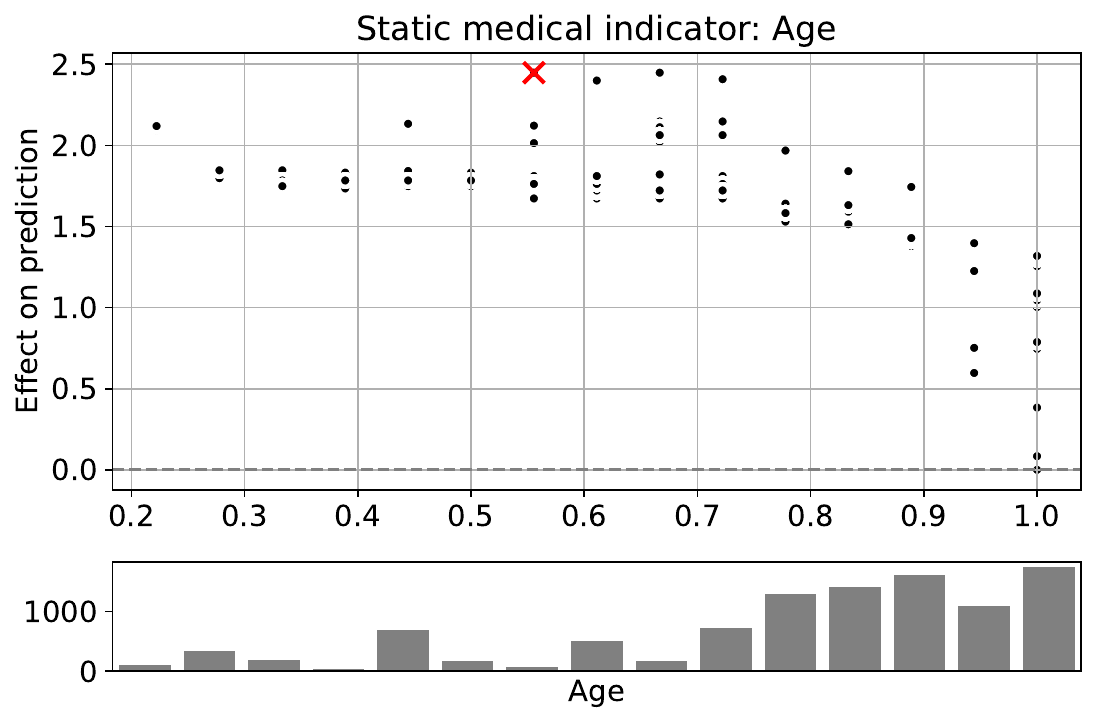}}
    \end{subfigure}
    \caption{Shape plot (left) and SHAP plot (right) for medical indicator \emph{Age}.}
    \label{fig:shape-vs-shap-age}
\end{figure}

\Cref*{fig:shape-vs-shap-age} shows two interpretation plots for the same static medical indicator \emph{Age} of our use case. While the medical indicator shape function retrieved from a PatWay-Net model with one interaction is illustrated on the left side, post-hoc generated \ac{shap} values for a black-box XGBoost model are shown on the right side. 
While the trend of both plots is similar, the \ac{shap} plot shows a high variance for the effect on the model prediction for a single value of the static medical indicator \emph{Age}. In contrast, the shape plot for the same indicator shows a continuous function that is presumably easier to comprehend for non-technical users.

\subsection{Runtime performance}
\label{sec:runtime}

Table \ref*{tab:runtime} compares the training and inference time of the static model of PatWay-Net's \ac{dnn} model, which is end-to-end trained, with a two-step version of the static module, which is not end-to-end trained. For the latter, we first trained an \ac{mlp} model per static feature and then used the output of all \ac{mlp} models as input for training and applying a subsequent logistic regression model. 
The experiments for this comparison were conducted on a workstation with 12 CPU cores and 128 GB RAM. 

The results show that the training time of PatWay-Net's static module with end-to-end training is on average 51.569 seconds, whereas the training time for the variant without end-to-end training takes on average 1.646,935 seconds. In other words, PatWay-Net's static module is about 33 times faster than training the individual models. Thus, the training of all \ac{mlp} models in one single architecture is considerably more efficient. 
Concerning inference, the average time for both variants is very low. 

\begin{table}[ht]
\footnotesize
\centering
\caption{Runtime performance.}
\label{tab:runtime}
\begin{tabular}{@{}lrr@{}}
\toprule
 & Training time (sec) & Inference time (sec) \\ \midrule
PatWay-Net (static module) & 51.569\textcolor{white}{1} ($\pm$30.790) & 0.005 ($\pm$.000) \\
MLPs + logistic regression & 1646.935 ($\pm$125.027) & 0.024 ($\pm$.001) \\ 
\bottomrule
\end{tabular}
\end{table}

\newpage
\section{Simulation study}
\label{app:simulationstudy}
In this appendix, we verify and demonstrate PatWay-Net's validity concerning the generated interpretation plots. More specifically, we follow the idea of other interpretable model proposals \citep[e.g.,][]{yang2021gami, agarwal2021neural}, in which the authors perform simulation studies based on synthetic data with controllable feature effects. In doing so, we prefer like in our real-life data application the designation \emph{(medical) indicator} over \emph{feature} because it is more comprehensible for decision-makers in the medical domain.

In the following, we provide details on the simulation data creation~(Appendix~\ref*{app:sim_data_creation}), the used experimental setting (Appendix~\ref*{app:sim_experimental_setting}), the obtained results~( Appendix~\ref*{app:sim_results}), and additional results (Appendix~\ref*{app:sim_additional_results}). 

\subsection{Simulation data creation}
\label{app:sim_data_creation} 

For our simulation study, we create a synthetic event log containing static and sequential medical indicators.\footnote{For the purpose of reproducibility, additional material can be found in the repository.}
The event log consists of 50,000 patient pathways with 12 events each. Every patient pathway begins with a start activity, called \emph{ER Registration}, followed by three measurements of the \emph{Heart Rate} and \emph{Blood Pressure} each, and then the administration of medication (four times \emph{A} and one time \emph{B} in random order). 
Further, the data set contains four static and two sequential medical indicators. Two of the static medical indicators are numerical, namely \emph{Age} and \emph{BMI} (Body Mass Index), whereas the remaining two are binary, namely \emph{Gender} and \emph{Foreigner}. They are all set randomly. The sequential medical indicator \emph{Heart Rate} occurs whenever the respective activity \emph{Heart Rate} appears. Thus, it has different values that either solely increase or decrease over time. In our simulation study, the increase or decrease is exemplarily always set to 30\%. 
The sequential medical indicator \emph{Blood Pressure} occurs whenever the respective activity \emph{Blood Pressure} appears. The values can increase and decrease randomly over time for one instance.

We created a continuous label for solving a regression-like prediction task. The label contains five different additive parts to introduce five different medical indicator effects concerning the static and sequential attributes in our event log: $y_{gender}$, $y_{age}$, $y_{pattern}$, $y_{hr-nl}$, and $y_{hr}$ (see \Cref*{equ:sim_label}). Each part can take values between 0 and 0.2, thus, the label $y$ can take values between 0 and 1. The remaining medical indicators (e.g., \emph{Foreigner} and \emph{BMI}) are only included as noise terms and do not have any effect on the target variable.
\begin{align}
\label{equ:sim_label}
 y &= y_{gender} + y_{age} + y_{pattern} + y_{hr-nl} + y_{hr}.
\end{align}
The first part $y_{gender}$ shows the influence of the static medical indicator \emph{gender}, where $b_{static}^{gender}$ is 1 if the patient is female and 0 otherwise:
\begin{align}
\label{equ:sim_label_gender}
 y_{gender} &= 0.2 * b_{static}^{gender}, \qquad \text{with } b_{static}^{gender}= \begin{cases}
    1,& \text{if } x_{static}^{gender}=1,\\
    0,              & \text{otherwise}.
\end{cases}
\end{align}
The second part, $y_{age}$, demonstrates the effect of the static medical indicator \emph{age} on the label, which we model as a downward open parabola:
\begin{align}
\label{equ:sim_label_age}
y_{age} &= \, -0.8 * ( x_{static}^{age}- 0.5)^2 + 0.2.
\end{align}
Besides the influence of static medical indicators on the label $y$, we also include an influence of sequential medical indicators. First, $b_{seq}^{pattern}$ is 1 if an instance contains a certain pattern in its sequence of activities regarding the administration of the medication, namely \emph{\textquote{Medication~A, Medication~A, Medication~A, Medication~A, Medication~B}}:
\begin{align}
\label{equ:sim_label_pattern}
y_{pattern} &= 0.2 * b_{seq}^{pattern}, \qquad \text{with } b_{seq}^{pattern} =
 \begin{cases}
    1,& \text{if } x_{t,seq}^{meda} = 1, \forall t \in \{8,\dots,11\} \\&\land \text{ } x_{12,seq}^{medb} = 1,  \\
    0,              & \text{otherwise}. \end{cases}
\end{align}
Further, $y_{hr-nl}$ shows the effect of the medical indicator \emph{Heart Rate} at time step $t_2$ as a downward open parabola:
\begin{align}
\label{equ:sim_label_hr}
y_{hr-nl} &= \, -0.8 * (x_{2,seq}^{hr} - 0.5)^2 + 0.2.
\end{align}
Lastly, $y_{hr}$ describes the effect of the behavior of the medical indicator \emph{Heart Rate} over time. The values can increase or decrease over time for a single patient pathway. If the values increase, $b_{seq}^{hr}$ will take the value 1, otherwise it will be 0:
\begin{align}
\label{equ:sim_label_hr_value}
y_{hr} &= 0.2 * b_{seq}^{hr}, \qquad \text{with } b_{seq}^{hr} =
 \begin{cases}
    1,& \text{if } x^{hr}_{t,seq}-x^{hr}_{t-1,seq} >0, \text{ for } t \in \{2,3,4\}, \\
    0,              & \text{otherwise}. \end{cases}
\end{align}

\subsection{Experimental setting}
\label{app:sim_experimental_setting}
We optimize PatWay-Net based on the complete simulation data set comprising 50,000 patient pathways and then generate interpretation plots based on 1,000 patient pathways of the same data set. 
Furthermore, we do not consider any sequential medical indicator interaction as we are interested in investigating how well the model can learn the five additive parts described in the previous section. 
Finally, we set the hidden size per sequential and static medical indicator to 16, and the batch size, number of epochs, and learning rate to 32, 1,000, and 0.001, respectively, as the loss converged well with these values. 

\subsection{Results}
\label{app:sim_results}
We present PatWay-Net's interpretation plots for the created synthetic event log and, based on the interpretation plots, we validate how well PatWay-Net captures the effects we modeled in the synthetic event log data.

\subsubsection{Importance of medical indicators}
\begin{figure}[ht]
    \centering
    \includegraphics[width=0.3\textwidth]{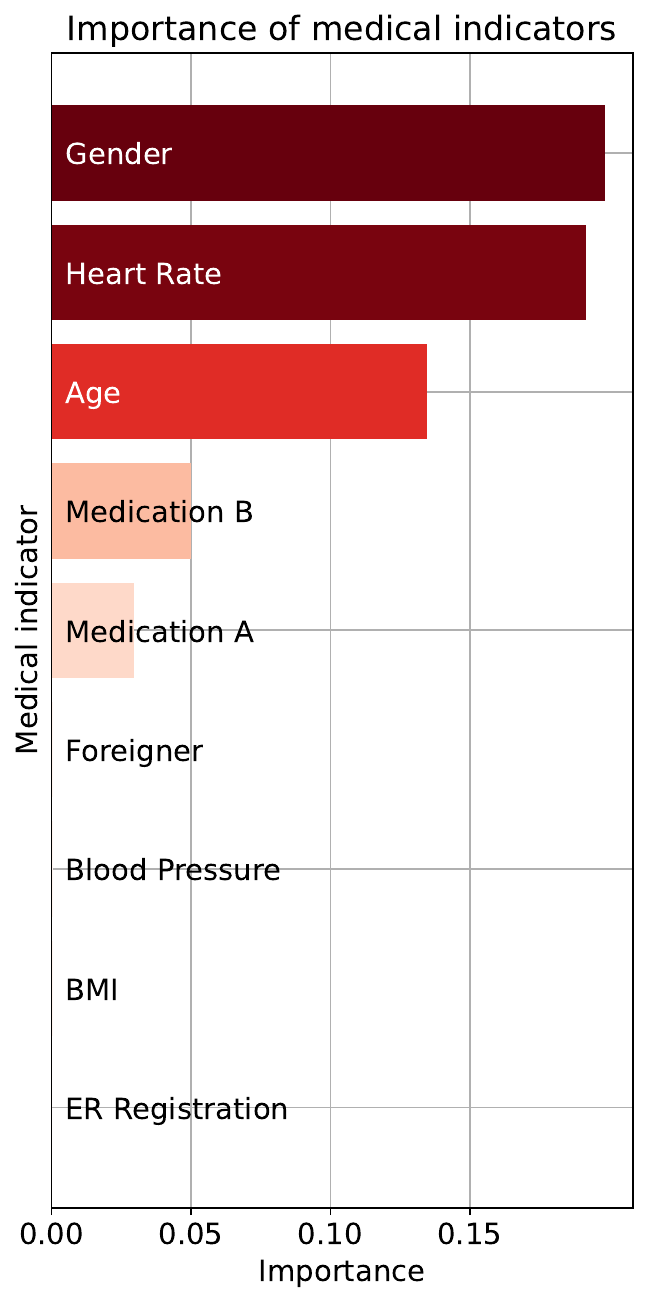}
    \caption{Importance for static and sequential medical indicators.}
    \label{fig:sim_globalfeatureimportance}
\end{figure}

\Cref*{fig:sim_globalfeatureimportance} shows the medical indicator importance plot at time step $t_{12}$. As expected, the medical indicators \emph{Gender}, \emph{Heart Rate}, \emph{Age}, \emph{Medication A}, and \emph{Medication B} show an effect on the model output. The remaining indicators correctly show no effect, as they were only included in the simulation to serve as irrelevant noise terms.

\subsubsection{Static medical indicator shape}
\begin{figure}[ht]
    \centering
    \begin{subfigure}{
        \includegraphics[height=3.7cm]{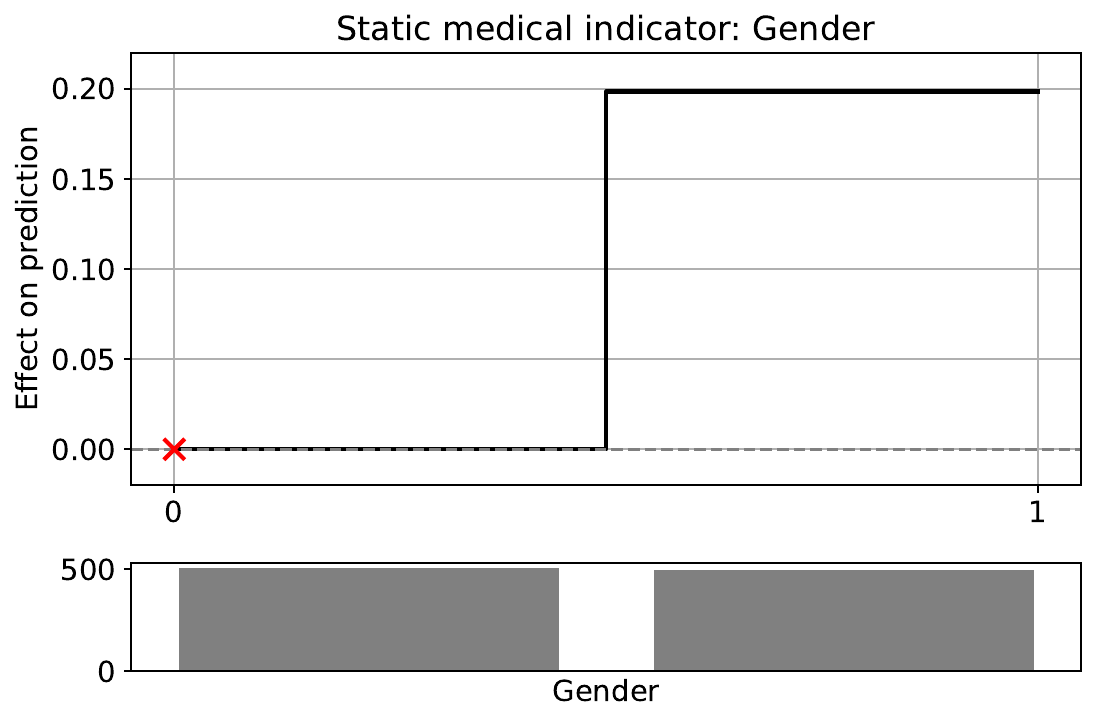}}
    \end{subfigure}%
    \begin{subfigure}{
        \includegraphics[height=3.7cm]{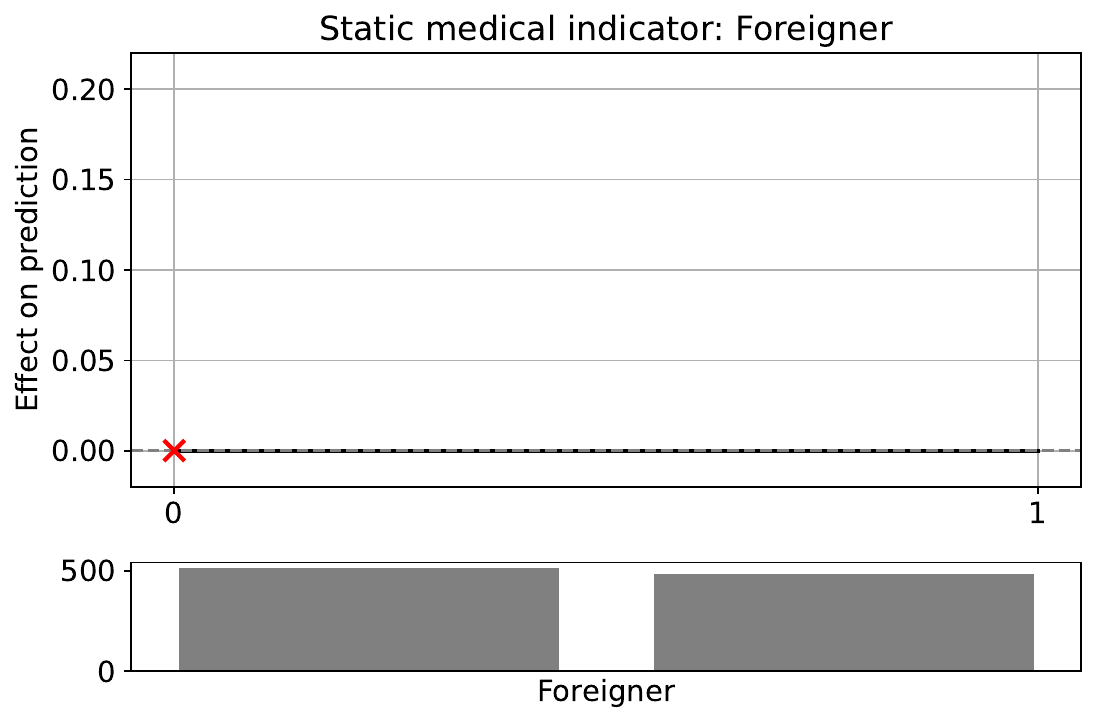}}
    \end{subfigure}
    \caption{Static medical indicator shape for indicators \emph{Gender} (left) and \emph{Foreigner} (right).}
    \label{fig:gender_foreigner}
\end{figure}

\Cref*{fig:gender_foreigner} compares the global effect of the model output for the static categorical medical indicators \emph{Gender} (left) and \emph{Foreigner} (right). PatWay-Net correctly detects the effect of the medical indicator \emph{Gender} with a constant value of 0.2, which equals the magnitude of the simulated coefficient whenever the indicator value is 1 (=~female). By contrast, the indicator \emph{Foreigner} has no effect.

\begin{figure}[ht]
    \centering
    \begin{subfigure}{
        \includegraphics[height=3.7cm]{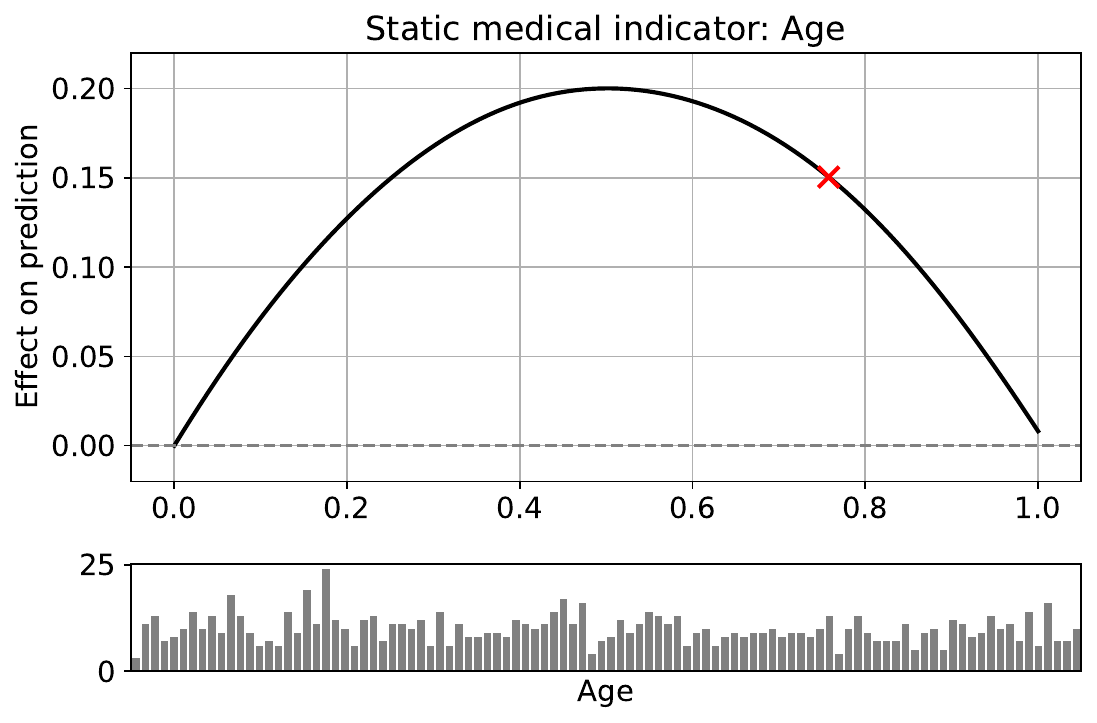}}
    \end{subfigure}%
    \begin{subfigure}{
         \includegraphics[height=3.7cm]{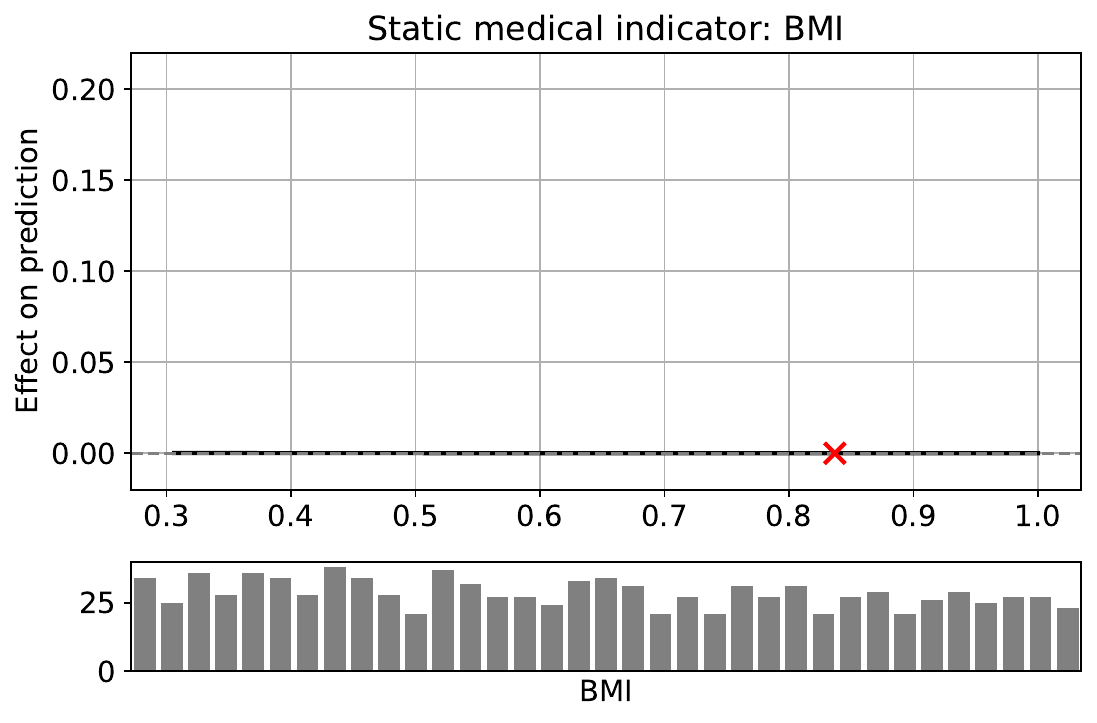}}
    \end{subfigure}
    \caption{Static medical indicator shape for indicators \emph{Age} (left) and \emph{BMI} (right).}
    \label{fig:age_bmi}
\end{figure}

Similarly, \Cref*{fig:age_bmi} shows the static medical indicator shape plot for indicator \emph{Age} (left) compared to the indicator \emph{BMI} (right).\footnote{For better interpretability, we applied post-processing to the effect values by adjusting the $y$-axis to 0.} 
PatWay-Net can correctly detect the parabolic effect on the label of the medical indicator \emph{Age}, with the strongest effect of 0.2, being achieved at a value of 0.5, as modeled in \Cref*{equ:sim_label_age}. By contrast, no influence was correctly detected for the indicator \emph{BMI}.

\subsubsection{Sequential medical indicator shape}

\begin{figure}[ht]
    \centering
    \begin{subfigure}{
        \includegraphics[height=3.7cm]{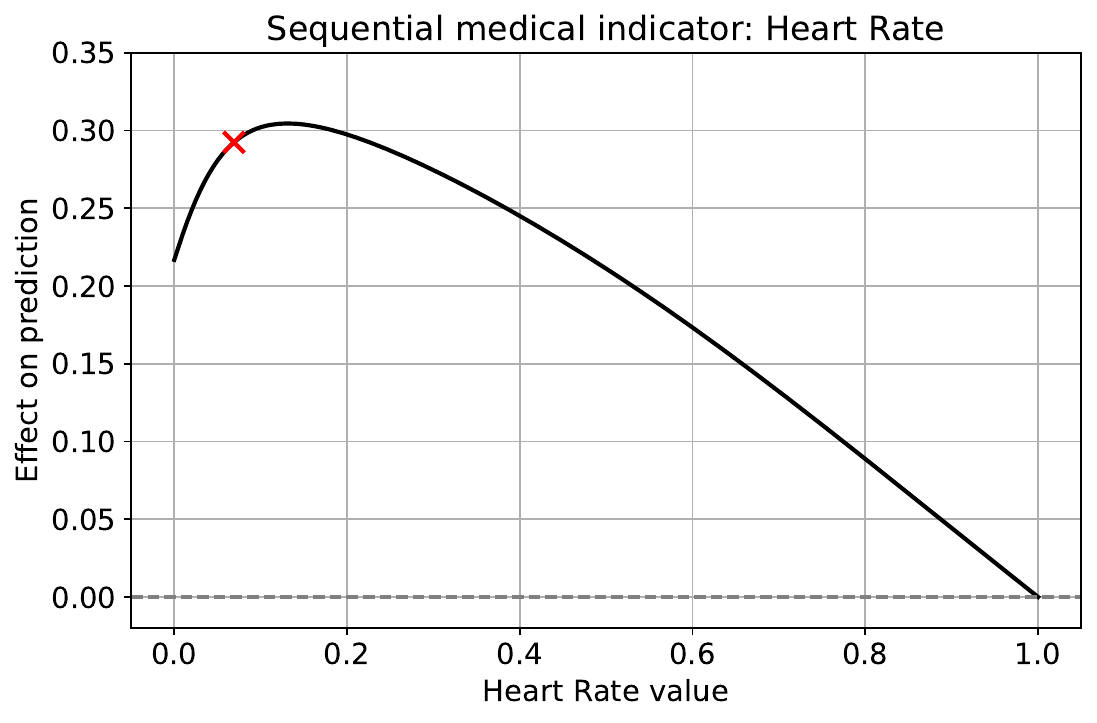}}
    \end{subfigure}%
    \begin{subfigure}{
        \includegraphics[height=3.7cm]{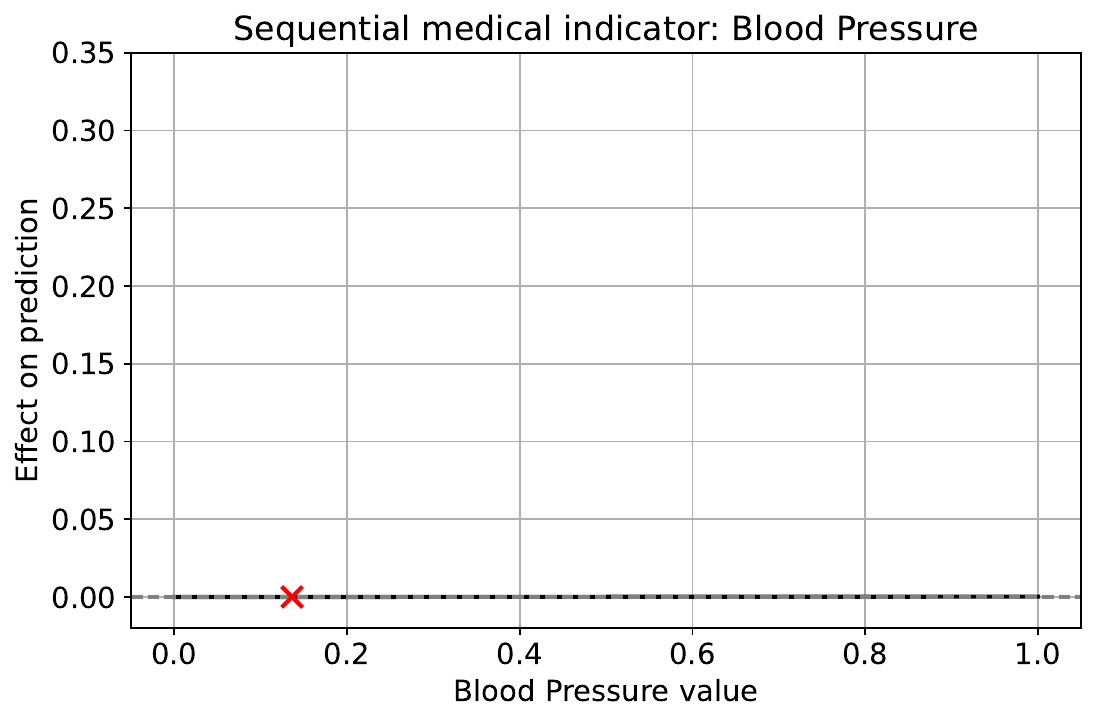 }}
    \end{subfigure}
    \caption{Sequential medical indicator shape for indicators \emph{Heart Rate} (left) and \emph{Blood Pressure} (right).}
    \label{fig:heartrate-bloodpres}
\end{figure}

For the sequential medical indicators \emph{Heart Rate} and \emph{Blood Pressure}, the indicator effect on the prediction can be evaluated for each time step. \Cref*{fig:heartrate-bloodpres} shows the sequential medical indicator shape for \emph{Heart Rate} (left) and \emph{Blood Pressure} (right) exemplarily for the time step $t_{4}$ and $t_{7}$, respectively.\footnote{We select these time steps because they are the last time steps at which a \emph{Heart Rate} and \emph{Blood Pressure} activity occur. However, the remaining figures can be found in the repository.} PatWay-Net can correctly detect the quadratic effect on the label for the indicator \emph{Heart Rate}. Indicator \emph{Blood Pressure}, on the other hand, has no effect, as correctly detected by PatWay-Net.

\subsubsection{Sequential medical indicator transition}

\begin{figure}[ht]
    \centering
    \begin{subfigure}{
        \includegraphics[height=3.9cm]{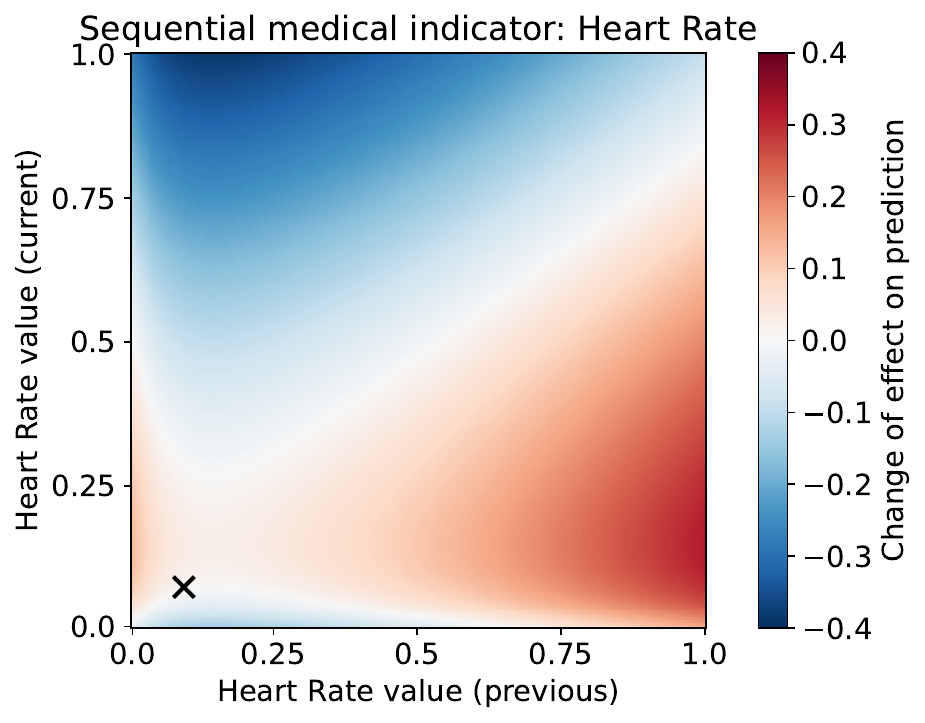}}
    \end{subfigure}%
    \begin{subfigure}{
        \includegraphics[height=3.9cm]{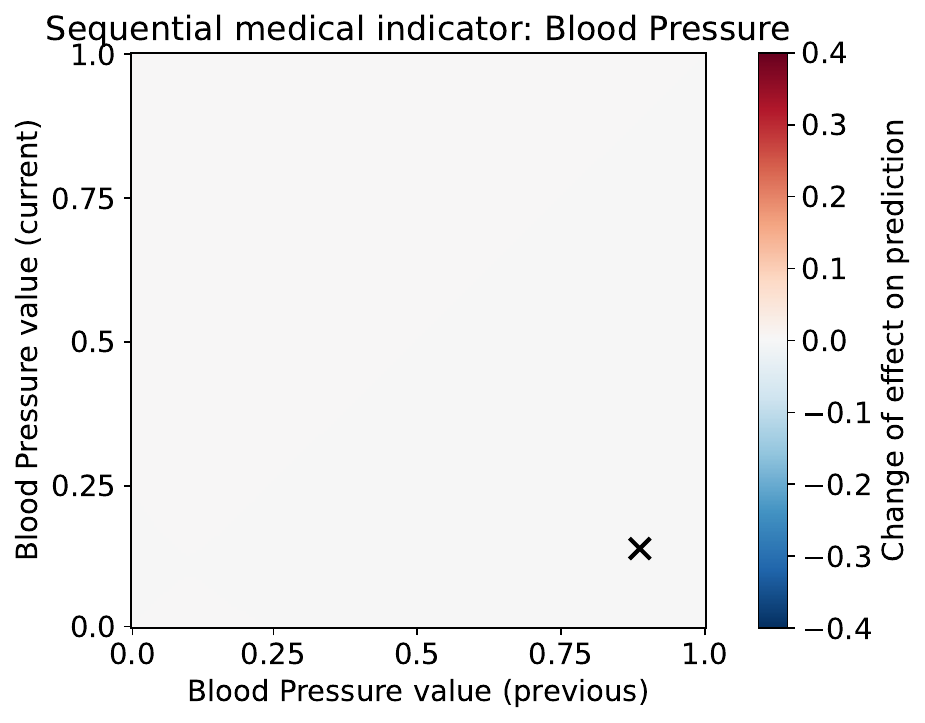}}
    \end{subfigure}
    \caption{
    Sequential medical indicator transition for indicators \emph{Heart Rate} (left) and \emph{Blood Pressure} (right).}
    \label{fig:sim_transition}
\end{figure}

As sequential medical indicators may change over time, it is crucial to investigate their effect not only at a certain time step but also the transition between two consecutive time steps. \Cref*{fig:sim_transition} shows exemplarily the sequential medical indicator transition of the indicator value as well as the change of effect on the prediction for the indicators \emph{Heart Rate} and \emph{Blood Pressure} from time step $t_3$ (previous) to time step $t_4$ (current). For the medical indicator \emph{Heart Rate}, the indicator value can either linearly increase or linearly decrease by 30\% from time step $t_3$ ($x$-axis) to $t_4$ ($y$-axis).
For increasing cases, we can observe the effect changes stronger negatively, that is, decreases from $t_3$ to $t_4$, from any indicator value to a higher indicator value (blue region). If the increase concerns two indicator values with a lower value, the change in the indicator effect is lower. This is due to the parabolic shape of the medical indicator \emph{Heart Rate} (see \Cref*{fig:heartrate-bloodpres}) as defined in \Cref*{equ:sim_label_hr_value}.
For the decreasing cases, we can observe the effect changes stronger positively, that is, increases from $t_3$ to $t_4$, from a higher to a lower indicator value (red region). As for increasing cases, if the decrease concerns two indicator values with a lower value, the change in the indicator effect is lower. Again, this is due to the parabolic shape of the medical indicator \emph{Heart Rate} (see~\Cref*{fig:heartrate-bloodpres}).

For the medical indicator \emph{Blood Pressure}, the indicator values can either randomly increase or decrease from time step $t_6$ ($x$-axis) to $t_7$ ($y$-axis), as demonstrated by the plot (see~\Cref*{fig:heartrate-bloodpres}). Further, the effect on the prediction of the medical indicator does not change over time, as the complete region is colored white, indicating a change in the indicator effect of~0. 
Thus, in summary, we can see that PatWay-Net correctly detects the effect of the transition of the \emph{Heart Rate} as modeled in \Cref*{equ:sim_label_hr_value}, whereas \emph{Blood Pressure} does not affect the model output.

\subsubsection{Sequential medical indicator development}
\begin{figure}[ht]
    \centering
    \includegraphics[width=0.80\textwidth]{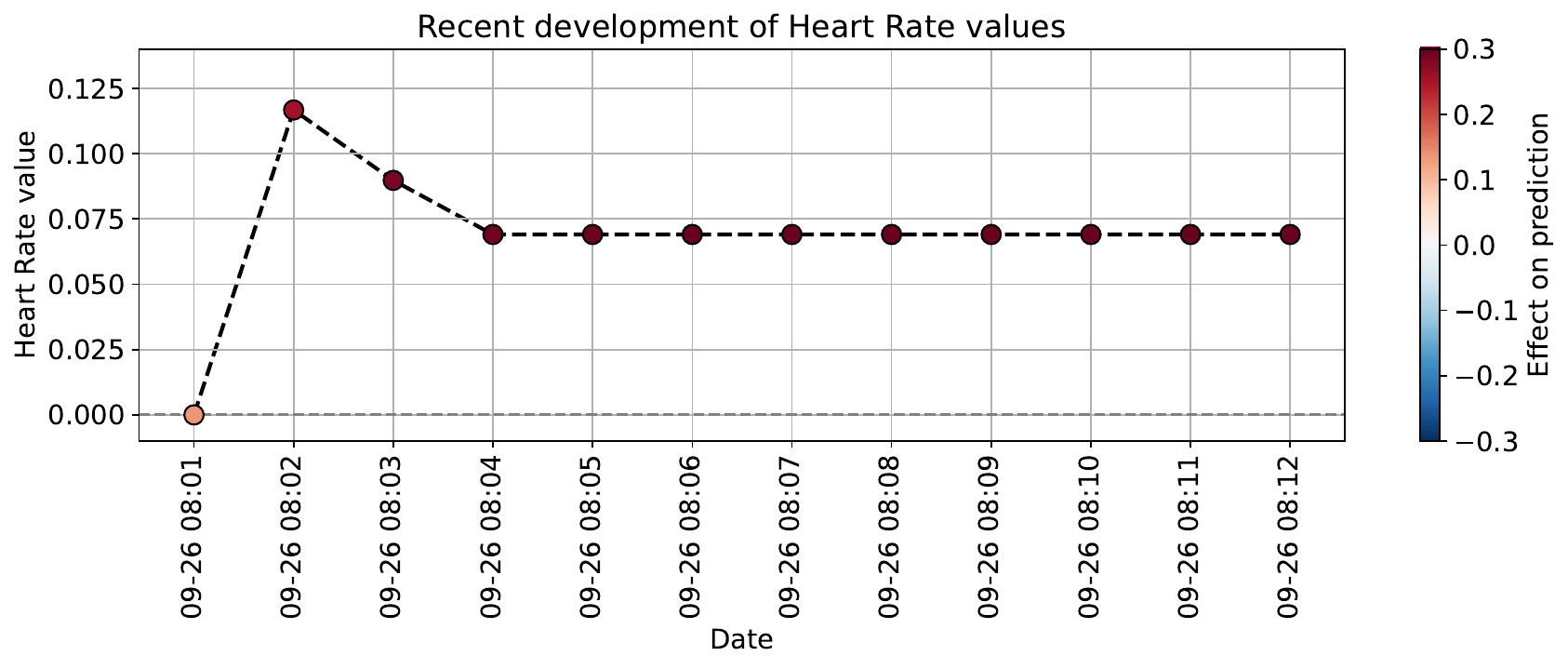}
    \caption{Sequential medical indicator development for indicator \emph{Heart Rate}.}
    \label{fig:sim_develop}
\end{figure}

\Cref*{fig:sim_develop} demonstrates the sequential medical indicator effect on the prediction for the indicator \emph{Heart Rate} for a given patient pathway over time. 
The strongest increase in the indicator effect can be observed from time step $t_1$ to $t_2$, as \emph{Heart Rate} is first measured at time step $t_2$. From $t_2$ to $t_3$, and from $t_3$ to $t_4$, the indicator effect increases more weakly, while the \emph{Heart Rate} value steadily decreases.  
After time step $t_4$, the effect remains the same, as the activity \emph{Heart Rate} does not occur after that. 

\subsection{Additional results}
\label{app:sim_additional_results}

We conduct additional experiments with the created synthetic event log and interpretable baselines regarding training performance and interpretability. 
\Cref*{tab:sim_perf} summarizes the average train performance of PatWay-Net models without an interaction and baseline models over five seeds. The train performance indicates how well \ac{ml} models capture the effects we modeled in the synthetic data. For measuring the training performance, we calculate the \ac{mse}, \ac{mae}, and $R^{2}$. As baselines, we use the interpretable shallow \ac{ml} approaches ridge regression, lasso regression, and decision tree regression. For training the lasso regression models and decision tree regression models, we set the regularization strength to 0.01 and the maximum depth to 3, respectively. We use default values for the remaining hyperparameters.
We observe that the PatWay-Net models considerably outperform the baseline models in terms of all regression measures. This indicates that the PatWay-Net models can better capture the modeled effects in the synthetic data.   

\begin{figure}[ht]
    \centering
    \includegraphics[width=0.7\textwidth]{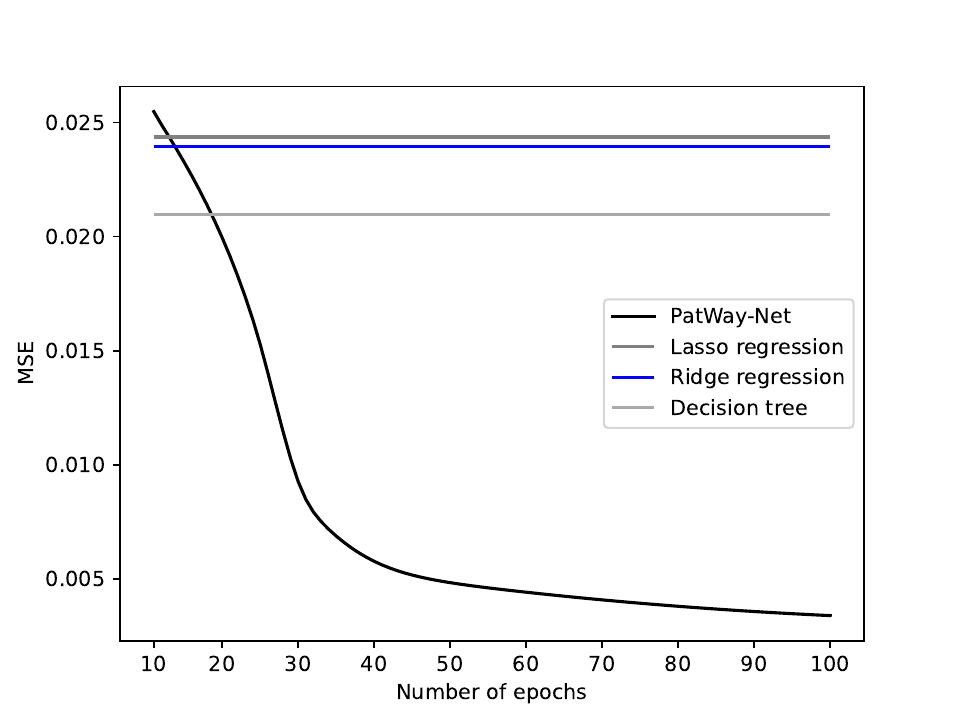}
    \caption{Loss of PatWay-Net (from epoch 10 to 100) and baselines.}
    \label{fig:loss}

\end{figure}

\Cref*{fig:loss} shows the loss curve for the training of PatWay-Net. The loss curve ranges from epoch 10 to 100 for better readability. Loss indicates the \ac{mse}. \ac{mse} values of the baseline models are added for a direct comparison.   
Based on this figure, the \ac{mse} values for the PatWay-Net model are increasingly lower from epoch 20 to 100. Therefore, the longer we train the PatWay-Net model, the better it can capture the modeled effects in the synthetic data.

\begin{table}[ht]
	\begin{center}
 \caption{Train performances for the simulation study.}
\label{tab:sim_perf}
\begin{tabular}{@{}lrrr@{}}
\toprule
          & Train \ac{mse} &Train \ac{mae}   &Train $R^{2}$    \\ \midrule
{PatWay-Net (without interaction)} &     0.000 \footnotesize{($\pm$.000)}                &    0.002 \footnotesize{($\pm$.000)}             &    0.997 \footnotesize{($\pm$.000)}                            \\
Ridge regression&  0.024 \footnotesize{($\pm$.000)}                 &    0.124 \footnotesize{($\pm$.000)}              &    0.290 \footnotesize{($\pm$.000)}                           \\
Lasso regression  &   0.024 \footnotesize{($\pm$.000)}                 & 0.126 \footnotesize{($\pm$.000)}                & 0.278 \footnotesize{($\pm$.000)}                           \\ 
Decision tree&0.021 \footnotesize{($\pm$.000)} &0.120 \footnotesize{($\pm$.000)}&0.379 \footnotesize{($\pm$.000)}\\
\bottomrule
\end{tabular}
\end{center}
\end{table}

\begin{table}[ht]
\centering
\caption{Model coefficients for static medical indicators of ridge regression.}
\label{tab:ridge_reg}
\begin{tabular}{lrrrr} \toprule
Static indicators & Gender     & Foreigner  & BMI         & Age        \\
\midrule
Model coefficients    & 0.197653 & 0.000666 & -0.001051 & 0.007274 \\\bottomrule
\end{tabular}
\end{table}

\begin{table}[ht]
\centering
\caption{Model coefficients for static medical indicators of lasso regression.}
\label{tab:lasso_reg}
\begin{tabular}{lrrrr} \toprule
Static indicators & Gender     & Foreigner  & BMI         & Age        \\
\midrule
Model coefficients    & 0.157634 & 0.00 & -0.00 & 0.00 \\
\bottomrule
\end{tabular}
\end{table}

\begin{figure}[ht]
    \centering
    \includegraphics[width=\textwidth]{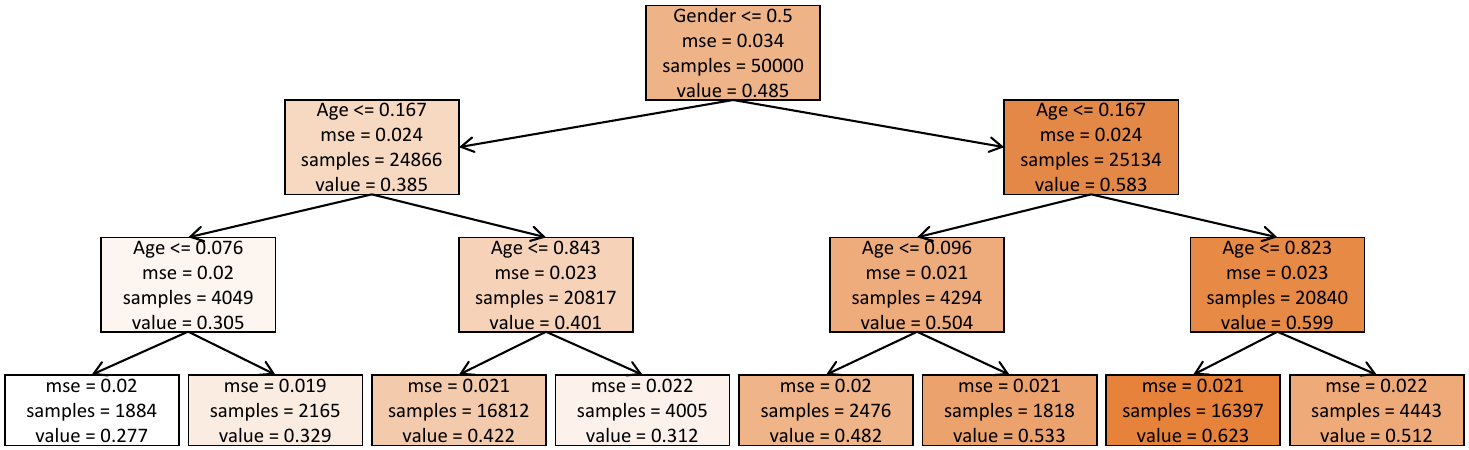}
    \caption{Decision tree model with depth 3.}
    \label{fig:decisiontree}
\end{figure}

\Cref*{tab:ridge_reg}, \Cref*{tab:lasso_reg}, and \Cref*{fig:decisiontree} represent coefficients of the ridge regression model, coefficients of the lasso regression model, and the structure of the decision tree regression model, respectively. 
By comparing the different interpretation plots of PatWay-Net presented in Appendix~\ref*{app:simulationstudy} with the outputs of the baseline models (i.e., model coefficients and tree structure), it becomes clear that PatWay-Net with its interpretation plots provides much more comprehensible model outputs than the interpretable shallow \ac{ml} models.

\newpage
\section{Expert interviews}
\label{sec:eval_interviews}

To assess PatWay-Net's interpretation quality, we conducted structured interviews with four independent medical experts. The medical experts are clinicians in different hospitals in different fields (surgery, trauma surgery, internal medicine, and anesthesiology), but are not directly associated with our use case. They have between five and ten years of professional experience.

The interviews were structured into five parts. First, all relevant information on the use case of patients with symptoms of sepsis was described to the interviewees. Second, a rough overview of different patient pathways was illustrated using a visualized process model. Third, general questions on predictive applications and their practical relevance were discussed. Fourth, we provided PatWay-Net's visual output and some additional descriptions of the interpretation in multiple steps. Based on this, we asked questions regarding usefulness, applicability, and trust in the predictions. Lastly, overarching questions on the dashboard were asked.\footnote{The interview guideline can be found in the online repository.} 

In general, the benefit of applying \ac{ml} models in the medical domain was confirmed by the interviewees, as \textquote{\textit{the idea of supporting medical differentiation through \ac{ml} makes a lot of sense}} \textbf{(I4)}. Specifically, using predictive approaches is helpful \textbf{(I2)} and \textquote{\textit{important for quick decision-making and provision or estimation of the required resources}} \textbf{(I1)}. It enables \textquote{\textit{better assessment of patient risk, prioritization, [...], and transfer to another hospital if necessary}} \textbf{(I4)}. Additionally, it fosters to \textquote{\textit{properly assess the patients}} \textbf{(I2)} and to \textquote{\textit{order closer monitoring of the blood values}} \textbf{(I3)}. Further, it enables to \textquote{\textit{better assess capacity}} \textbf{(I2)}, so that practitioners \textquote{\textit{can also adjust \ac{icu} beds}} \textbf{(I3)}. 
In particular, the admission to \ac{icu} is helpful for medical experts with less experience: \textquote{\textit{For younger colleagues [...] who don't have that much experience with [determination of \ac{icu} admission] yet, it's certainly helpful, also just to make sure that nothing is overlooked}}~\textbf{(I2)}.

However, all interviewees confirmed that they \textquote{\textit{would not trust such predictions without further explanation}} \textbf{(I1)} as they \textquote{\textit{think it's important to know what the decision is based on}} \textbf{(I2)}. Clinicians \textquote{\textit{have to justify the process to the patients and relatives and to do that [they] want to understand the process to be able to justify it}} \textbf{(I4)}.

Our dashboard was evaluated to be \textquote{\textit{helpful as a support for decision-making}} \textbf{(I1)}, because \textquote{\textit{it is similar in processes to [their] own decision-making}} \textbf{(I1)} and it is \textquote{\textit{prepared very well}} \textbf{(I3)}. Additionally, the interviewees \textquote{\textit{believe these tools will not miss findings and will guide you to look at everything again when it automatically shows up in an overview}} \textbf{(I1)}. 

Moreover, all medical experts confirm the usefulness of the interpretation plots as, for example, \textbf{I2} \textquote{\textit{find[s] it helpful in any case [to] have an overview of which factors have been included [and to] [...] understand at a glance what caused the system to do this. [...] Because then [they] can also take a closer look at what the blood pressure is doing or where the leukocytes are}}. They positively assess the visual plots and think that \textquote{\textit{this is the language that is spoken medically in actually every continent}} \textbf{(I3)}. All interviewees stated that they prefer simple plots, because \textquote{\textit{there are always so many new colleagues [...] in the clinic that I think keeping it as simple as possible makes the most sense}} \textbf{(I2)} and that they usually have to act \textquote{\textit{relatively quickly}} \textbf{(I2, I3)}.

They prefer the suggested plots over common \ac{shap} plots because they look \textquote{\textit{clearer and [are] easier to follow}} \textbf{(I2, I4)}. They feel that \ac{shap} plots are \textquote{\textit{confusing and not really self-explaining}} because there are \textquote{\textit{too many dots in the plot}} \textbf{(I3)} (cf. also Appendix~\ref*{app:use_case}). Finally, one interviewee suggested providing interpretations in the form of \textquote{\textit{written text}} \textbf{(I4)}.

All interviewees also confirm that interpretations in the form of PatWay-Net's dashboard \textquote{\textit{would positively influence [their] [...] trust in the predictions}} \textbf{(I1, I4)}, because the dashboard provides \textquote{\textit{exactly what [they] work out [themselves] during a [...] visit}} \textbf{(I3)}. Therefore, they think that it \textquote{\textit{increases [the] acceptance of such predictions}} \textbf{(I3)}.

\newpage
\section{Additional use cases}
\label{app:furtherdata}
To prove PatWay-Net's robustness and generalizability, we use two additional data sets from different domains (i.e., hospital billing and loan application). The event logs are publicly available, and we use the version by Teinemaa et al.~\citep{teinemaa2019outcome} as they include labels for process outcome prediction. The data sets are preprocessed in the same way as for the event log (Appendix~\ref*{app:preprocessing}). To keep the computational effort manageable, we only used the first 5,000 traces. Interactions of the models are detected and incorporated as for the use case (Appendix~\ref*{app:automatic_search_interactions}). The hyperparameters used for the experiments can be found in the repository. 

First, we use the \emph{hospital billing}\footnote{\url{https://research.tue.nl/en/datasets/hospital-billing-event-log}} event log. 
It captures billing data from an \ac{erp} system of a hospital about conducted services. Its preprocessed version contains two static medical indicators. These two indicators, namely \emph{speciality} and \emph{caseType}, are categorical and contain 22 and eight different values, respectively. Besides the activity with 18 different values, the preprocessed event log contains another sequential medical indicator, namely \emph{state}, which is also categorical with ten different values. Finally, the label indicates whether a case is reopened \citep{teinemaa2019outcome}.
The second data set is the \emph{bpi2012}\footnote{\url{https://www.win.tue.nl/bpi/doku.php?id=2012:challenge}} event log. The data was taken from a Dutch financial institute and captures the process of loan applications. After preprocessing, it contains one static numerical medical indicator, namely \emph{amount\_req}, and the activity includes 36 different values. The label indicates if the application was accepted.

The results for PatWay-Net and the baselines for both data sets are summarized in \Cref*{tab:results_hospitalbilling} and \Cref*{tab:results_bpi2012}.
For both event logs, we can show that PatWay-Net outperforms the baselines regarding the $AUC_{ROC}$. 
For the bpi2012 event log, we observe that PatWay-Net also outperforms the baselines regarding the F1-score.

\begin{table}[ht]
\centering
\caption{Comparison between baseline models and PatWay-Net for the hospital billing event log.}
\label{tab:results_hospitalbilling}
\resizebox{\textwidth}{!}{
\begin{tabular}{@{}lrrlrr@{}}
\toprule
\begin{tabular}{l}\multirow{2}{*}{\ac{ml} approach}\end{tabular} & \multicolumn{2}{c}{F1-score (weighted)}                   &  & \multicolumn{2}{c}{$AUC_{ROC}$}                               \\ \cmidrule(lr){2-3} \cmidrule(l){5-6} 
                          & \multicolumn{1}{c}{Validation} & \multicolumn{1}{c}{Test} &  & \multicolumn{1}{c}{Validation} & \multicolumn{1}{c}{Test} \\ \midrule
\multicolumn{6}{l}{\footnotesize{\textsc{Our Approach}}}                                                                 \\
\begin{tabular}{l}PatWay-Net\\(with interaction)\end{tabular}                &  0.944 \footnotesize{($\pm$.006)}       & {0.950} \footnotesize{($\pm$.004)} &  &  0.691 \footnotesize{($\pm$.049)}    & {0.627} \footnotesize{($\pm$.040)} \\
\begin{tabular}{l}PatWay-Net\\(without interaction)\end{tabular}                &  0.944 \footnotesize{($\pm$.006)}       &  0.950 \footnotesize{($\pm$.004)} &  &  0.694 \footnotesize{($\pm$.053)}       & 0.614 \footnotesize{($\pm$.056)} \\
\multicolumn{6}{l}{\footnotesize{\textsc{Interpretable Shallow Machine Learning}}}                                                                                           \\
\begin{tabular}{l}Decision tree  \end{tabular}                      &      0.944 \footnotesize{($\pm$.006)}                             &      0.950 \footnotesize{($\pm$.005)}                       &  &             0.589 \footnotesize{($\pm$.071)}                      &    0.546 \footnotesize{($\pm$.043)}                        \\
\begin{tabular}{l}$K$-nearest neighbor  \end{tabular}                     &    0.932 \footnotesize{($\pm$.010)}                             &      0.936 \footnotesize{($\pm$.012)}                     &  & 0.512 \footnotesize{($\pm$.020)}                                &        0.534 \footnotesize{($\pm$.023)}                   \\
\begin{tabular}{l}Na\"ive Bayes    \end{tabular}                    &  0.942 \footnotesize{($\pm$.004)}                               &       0.947 \footnotesize{($\pm$.010)}                    &  &      0.670 \footnotesize{($\pm$.051)}                           &          0.605 \footnotesize{($\pm$.048)}                 \\
\begin{tabular}{l}Logistic regression  \end{tabular}                      &   0.944 \footnotesize{($\pm$.006)}                              &    0.950 \footnotesize{($\pm$.004)}                       &  &    0.674 \footnotesize{($\pm$.055)}                             &      0.599 \footnotesize{($\pm$.049)}                     \\
\multicolumn{6}{l}{\footnotesize{\textsc{Non-Interpretable Machine Learning}}}\\

\begin{tabular}{l}\ac{lstm} network\\(with static module) \end{tabular}                      &    0.966 \footnotesize{($\pm$.014)}                            &   0.964 \footnotesize{($\pm$.007)}                       &  &        0.814 \footnotesize{($\pm$.057)}                        &   0.758 \footnotesize{($\pm$.103)}                       \\
\begin{tabular}{l}XGBoost \end{tabular}                       &   0.944 \footnotesize{($\pm$.006)}                              &    0.951 \footnotesize{($\pm$.005)}                       &  &    0.696 \footnotesize{($\pm$.054)}                             &      0.623 \footnotesize{($\pm$.038)}                     \\
\begin{tabular}{l}Random forest \end{tabular}                       &   0.944 \footnotesize{($\pm$.006)}                              &    0.951 \footnotesize{($\pm$.005)}                       &  &    0.692 \footnotesize{($\pm$.060)}                             &      0.626 \footnotesize{($\pm$.047)}                     \\
\bottomrule
\end{tabular}
}
\end{table}

\begin{table}[ht]
\caption{Comparison between baseline models and PatWay-Net for the bpi2012 event log.}
\label{tab:results_bpi2012}
\centering
\resizebox{\textwidth}{!}{
\begin{tabular}{@{}lrrlrr@{}}
\toprule
\begin{tabular}{l}\multirow{2}{*}{\ac{ml} approach}\end{tabular} & \multicolumn{2}{c}{F1-score (weighted)}                   &  & \multicolumn{2}{c}{$AUC_{ROC}$}                               \\ \cmidrule(lr){2-3} \cmidrule(l){5-6} 
                          & \multicolumn{1}{c}{Validation} & \multicolumn{1}{c}{Test} &  & \multicolumn{1}{c}{Validation} & \multicolumn{1}{c}{Test} \\ \midrule
\multicolumn{6}{l}{\footnotesize{\textsc{Our Approach}}}                                                                 \\
\begin{tabular}{l}PatWay-Net\\(with interaction)\end{tabular}               &  0.638 \footnotesize{($\pm$.008)}       & {0.662} \footnotesize{($\pm$.003)} &  &  0.740 \footnotesize{($\pm$.006)}       & {0.750} \footnotesize{($\pm$.003)} \\
\begin{tabular}{l}PatWay-Net\\(without interaction)\end{tabular}               &  0.647 \footnotesize{($\pm$.006)}       &  0.665 \footnotesize{($\pm$.009)} &  &  0.741 \footnotesize{($\pm$.007)}       & 0.750 \footnotesize{($\pm$.005)} \\
\multicolumn{6}{l}{\footnotesize{\textsc{Interpretable Shallow Machine Learning}}}                                                                                           \\
\begin{tabular}{l}Decision tree   \end{tabular}                     &      0.319 \footnotesize{($\pm$.012)}                             &      0.369 \footnotesize{($\pm$.002)}                       &  &             0.519 \footnotesize{($\pm$.005)}                      &    0.529 \footnotesize{($\pm$.007)}                        \\
\begin{tabular}{l}$K$-nearest neighbor  \end{tabular}                     &    0.512 \footnotesize{($\pm$.026)}                             &      0.527 \footnotesize{($\pm$.019)}                     &  & 0.528 \footnotesize{($\pm$.024)}                                &        0.534 \footnotesize{($\pm$.021)}                   \\
\begin{tabular}{l}Na\"ive Bayes    \end{tabular}                    &  0.412 \footnotesize{($\pm$.075)}                               &       0.458 \footnotesize{($\pm$.072)}                    &  &      0.515 \footnotesize{($\pm$.014)}                           &          0.542 \footnotesize{($\pm$.029)}                 \\
\begin{tabular}{l}Logistic regression \end{tabular}                       &   0.320 \footnotesize{($\pm$.012)}                              &    0.369 \footnotesize{($\pm$.002)}                       &  &    0.473 \footnotesize{($\pm$.013)}                             &      0.490 \footnotesize{($\pm$.016)}                     \\
\multicolumn{6}{l}{\footnotesize{\textsc{Non-Interpretable Machine Learning}}}\\

\begin{tabular}{l}\ac{lstm} network\\(with static module) \end{tabular}                     &    0.648 \footnotesize{($\pm$.014)}                            &   0.672 \footnotesize{($\pm$.009)}                       &  &        0.744 \footnotesize{($\pm$.008)}                        &   0.760 \footnotesize{($\pm$.012)}                       \\
\begin{tabular}{l}XGBoost \end{tabular}                       &   0.448 \footnotesize{($\pm$.027)}                              &    0.509 \footnotesize{($\pm$.032)}                       &  &    0.520 \footnotesize{($\pm$.003)}                             &      0.554 \footnotesize{($\pm$.013)}                     \\
\begin{tabular}{l}Random forest \end{tabular}                       &   0.352 \footnotesize{($\pm$.040)}                              &    0.414 \footnotesize{($\pm$.065)}                       &  &    0.521 \footnotesize{($\pm$.010)}                             &      0.537 \footnotesize{($\pm$.012)}                     \\

\bottomrule
\end{tabular}
}
\end{table}

\end{appendices}

\clearpage
\newpage
\bibliography{sn-bibliography}

\end{document}

%% file: acronyms.tex
\DeclareAcronym{shap}{
	short = SHAP,
	long = Shapley additive explanations
}

\DeclareAcronym{xgb}{
	short = XGB,
	long = XGBoost
}

\DeclareAcronym{lime}{
	short = LIME,
	long = local interpretable model-agnostic explanations
}

\DeclareAcronym{xai}{
	short = XAI,
	long = explainable artificial intelligence
}

\DeclareAcronym{erp}{
	short = ERP,
	long = enterprise resource planning
}

\DeclareAcronym{mlp}{
	short = MLP,
	long = multi-layer perceptron
}

\DeclareAcronym{ml}{
	short = ML,
	long = machine learning
}

\DeclareAcronym{dnn}{
	short = DNN,
	long = deep neural network
}

\DeclareAcronym{mae}{
	short = MAE,
	long = mean absolute error
}

\DeclareAcronym{mse}{
	short = MSE,
	long = mean squared error
}

\DeclareAcronym{lstm}{
	short = LSTM,
	long = long short-term memory
}

\DeclareAcronym{ilstm}{
    short = iLSTM,
    long = interpretable LSTM,
}

\DeclareAcronym{icu}{
    short = ICU,
    long = intensive care unit,
}

\DeclareAcronym{ncu}{
    short = NCU,
    long = normal care unit,
}

\DeclareAcronym{er}{
    short = ER,
    long = emergency room,
}

\DeclareAcronym{gam}{
    short = GAM,
    long = generalized additive model,
}

\DeclareAcronym{ebm}{
    short = EBM,
    long = explainable boosting machine,
}

\DeclareAcronym{gbm}{
    short = GBM,
    long = gradient boosting machine,
}

\DeclareAcronym{gaim}{
    short = GAIM,
    long = generalized additive index model,
}

\DeclareAcronym{nam}{
    short = NAM,
    long = neural additive model,
}

\DeclareAcronym{crp}{
    short = CRP,
    long = C-reactive protein,
}